\documentclass[preprint,12pt, 3p, titlepage]{elsarticle}

\usepackage{latexsym}
\usepackage{amssymb}
\usepackage{amsmath}  
\usepackage{amsthm}
\usepackage{booktabs}
\usepackage{enumitem}  
\usepackage{graphicx}
\usepackage{color}
\usepackage[ruled,algonl,vlined]{algorithm2e}
\usepackage{xcolor}
\usepackage{adjustbox}
\usepackage[utf8]{inputenc}
\usepackage{mathtools}  
\usepackage{multirow}
\usepackage{caption,subcaption}
\usepackage{url}
\usepackage{float}
\usepackage{soul}
\usepackage{setspace}
\usepackage{fixmath}
\usepackage{lineno}

\newtheorem{theorem}{Theorem}

\newtheorem{proposition}[theorem]{Proposition}

\newtheorem{definition}{Definition}

\newcommand{\BibTeX}{B\kern-.05em{\sc i\kern-.025em b}\kern-.08em\TeX}

\begin{document}


\begin{frontmatter}

\title{Geometric Preference Elicitation for Minimax Regret Optimization in Uncertainty Matroids}

\author[1]{Aditya Sai Ellendula}
\ead{aditya20uari005@mahindrauniversity.edu.in}
\author[1]{Arun K Pujari}
\ead{arun.k.pujari@gmail.com}
\author[2]{Vikas Kumar\corref{cor1}}
\ead{vikas@cs.du.ac.in}
\author[3]{Venkateswara Rao Kagita}
\ead{venkat.kagita@nitw.ac.in}

\address[1]{Mahindra University, Hyderabad, India} 
\address[2]{University of Delhi, New Delhi, India}
\address[3]{National Institute of Technology, Warangal, India}
    
\cortext[cor1]{Corresponding author}

\begin{abstract}
This paper presents an efficient preference elicitation framework for uncertain matroid optimization, where precise weight information is unavailable, but insights into possible weight values are accessible. The core innovation of our approach lies in its ability to systematically elicit user preferences, aligning the optimization process more closely with decision-makers' objectives. By incrementally querying preferences between pairs of elements, we iteratively refine the parametric uncertainty regions, leveraging the structural properties of matroids. Our method aims to achieve the exact optimum by reducing regret with a few elicitation rounds.  Additionally, our approach avoids the computation of Minimax Regret and the use of Linear programming solvers at every iteration, unlike previous methods. Experimental results on four standard matroids demonstrate that our method reaches optimality more quickly and with fewer preference queries than existing techniques.

\end{abstract}


\end{frontmatter}


\section{Introduction}
\label{Sec:Intro}
The aim of a decision support system is to assist users in decision-making, with a focus on decision analysis and artificial intelligence in the development of automated decision support process. Keeping in mind the diversity of interests of the users,  these systems need to accommodate variations in user preferences, often utilizing methods such as inference from observed behavior. In this context, preference elicitation (PE) \cite{toffano2020efficient} is a fundamental problem in the development of intelligent decision tools and autonomous agents.  As a strategic approach to gathering information, PE has long been employed to assess users' preferences across various domains, including decision analysis \cite{KeeneyRaiffa1976, Salo2001}, marketing science \cite{toubia2004}, and AI \cite{chajewska2000making, freedman2020adapting, boutilier2002pomdp}. When initial information is lacking, preference elicitation techniques strive to gather user preference information to enable the systems to assist in achieving the objectives \cite{wang2003incremental}.  Various approaches to this problem have been proposed \cite{chajewska2000making, boutilier2002pomdp, white1984model, benabbou2021interactive, bourdache2019active}.

The problem of determining the maximum or minimum weight base in a matroid with perfect knowledge regarding weights is a well-studied optimization problem. When situations involve uncertain weights, the Minimax Regret (MMR) criterion is particularly suitable. Notably, the MMR criterion has been extensively studied for various combinatorial problems \cite{candia2011minmax}, including spanning trees and shortest paths \cite{kasperski2012tabu, montemanni2004branch}. Solving optimization problems with the MMR criterion is NP-hard, and researchers try to propose approximation algorithms \cite{kouvelis2013robust, boutilier2006constraint, hyafil2006regret, renou2010minimax, regan2012regret, baak2023generalizing}. Alternatively, researchers have proposed elicitation processes as a means to mitigate intractability by incorporating additional knowledge until the problem can be solved, either exactly or approximately \cite{white1984model, bourdache2019active}. Most work on elicitation assumes some uncertainty representation over user preferences or utility functions and uses one or more types of query to reduce that uncertainty. A variety of work uses strict uncertainty in which the parameters of a user’s utility function are assumed to lie in some region (e.g., polytope), and queries are used to refine this region. With minimax regret criterion, queries can be selected based on their ability to reduce the volume or surface of a polytope, to decrease minimax regret.

In this study, we address the challenge of determining the optimal weight-sum base of a matroid under uncertain weights by integrating a preference elicitation (PE) process. Rather than adopting traditional methods that model uncertainty as the Cartesian product of uncertain elements, we represent uncertainty as a parametric polyhedral region in the parameter space \cite{benabbou2021combining}. The core of our approach is an adaptive, online PE strategy, where we systematically solicit user preferences between pairs of elements to iteratively refine the polyhedral uncertainty region by characterizing extreme points (and their adjacencies) of polyhedron bounding the uncertainty region instead of using an LP solver to explore the uncertainty set. 

In this study, we introduce an algorithm that skillfully employs matroidal properties, offering an efficient preference elicitation to the aforementioned optimization problem. Our approach selects queries to reduce the number of rounds. The key aspects of our approach are detailed below:

\begin{itemize}
    \item In each iteration, we progressively refine the polyhedral uncertainty region by leveraging polyhedral combinatorics to generate extreme points and their adjacencies. Numerical computation is employed selectively, calculating coordinates exclusively for new points, while combinatorial methods generate their adjacencies. This stands in contrast to earlier approaches which use external LP solvers to handle this task.
    \item We propose a conflict-directed heuristic to generate preference queries with the objective of  decreasing the MMR value faster.
    \item The proposed algorithm estimates an upper bound of MMR and avoids calculating MMR at each iteration. This bound enables the algorithm to detect when MMR reaches zero. For non-zero MMR termination criteria, the heuristic remains useful, streamlining the process compared to previous proposals  \cite{benabbou2021combining} that require MMR computation in every iteration.
    \item Our experimental analysis across four distinct matroids shows that our algorithm attains the desired solution with a reduced number of elicitation rounds, all while maintaining computational efficiency. 
    \item The proposed strategy leads to significant computational efficiency gains. 
\end{itemize}

The paper's organization is as follows: Section~\ref{Sec:Background} provides background on uncertainty matroids, minimax regret (MMR), and preference elicitation. In Section~\ref{Sec:Pr}, we formally state the problem, review prior research, and highlight our contributions. Section~\ref{Sec:PolyCompApproach} outlines the key components of our proposed method. The algorithm is presented in Section~\ref{Sec:Pr-A}. Experimental analyses, comparing algorithms and demonstrating the advantages of our approach are detailed in Section~\ref{Sec:Exp}. Finally, conclusions are drawn in Section~\ref{Sec:Con}.

\section{Background}
\label{Sec:Background}
This section introduces fundamental concepts in matroid theory and the concept of uncertainty matroids.  It also introduces the minimax regret criterion for optimization.  Furthermore, we introduce the concept of preference elicitation, which plays a significant role in shaping our theoretical framework. 

\subsection{Classical Optimization Problem on Matroid}

Let $E=\{e_1,e_2,\ldots,e_n\}$ be a finite set with $|E|=n$. A matroid $M$ is defined as $M=(E,I)$ with $E$ as the ground set together with a collection of sets $I\in 2^E$ known as the independent sets, satisfying the following axioms \cite{Pitsoulis2014, welsh1976}:

\begin{itemize}
    \item  If $S\in I$ and $T\subseteq S$, then $T\in I$.
    \item  If $S\in I$, $T\in I$, and $|T|>|S|$, then there exists $e\in T\setminus S$ such that $S\cup\{e\}\in I$.
\end{itemize}

A maximal independent set of a matroid is referred to as a base. All bases of a matroid have the same cardinality, and is the rank of the matroid $r(M)$. The set of all bases is denoted as $\mathcal{B}$. Various types of matroids can be defined based on specific structures or properties. 

\vspace{0.1cm}
\noindent
\textbf{Graphic Matroid}: A {\em graphic matroid} is defined with $E$ as the set of edges of a connected graph $G$ by taking the edges of the various spanning trees of $G$ as bases of the matroid.   

\vspace{0.1cm}
\noindent
\textbf{Uniform Matroid}: The $k$-{\em uniform matroid}  on $E$ is obtained by taking as bases all subsets of $E$ with exactly $k$  elements. 

\vspace{0.1cm}
\noindent
\textbf{Scheduling Matroid}: A {\em scheduling matroid}  has a set  of jobs as its ground set.  Each job requires one unit of processing time and has an integer release time and deadline.  A subset of jobs is independent if and only if there is a feasible schedule for them.

\vspace{0.1cm}
\noindent
\textbf{Partition Matroid}: A matroid $M$ is a {\em partition matroid} if the ground set $E$  can be partitioned into disjoint sets $P_1, P_2 \ldots, P_d$ such that $S \subseteq E$ is independent if and only if $|S \cap P_i| \leq t$ for all $1 \leq i \leq d$. We take $t$ to be 1 here.

In a weighted matroid, a nonnegative, additive weight $w_e$ is associated with every element $e\in E$ and the weight of any subset $S$ is $w(S)=\sum_{e\in S} w_e$. We denote $W=(w_e)_{e\in E}$ as the vector of all weights. The problem of finding a maximum or minimum weight base of a matroid \cite{edmonds1971matroids} is well-established combinatorial optimization problem as follows.

\begin{equation}
\text{Minimize}_{B\in \mathcal{B}} \sum_{e\in B} w_e \tag{P1} \label{eq:P1}
\end{equation}

 We define $z^*=\min_{B\in \mathcal{B}} w(B)$ as the optimal weight and $B^*=\arg\min_{B\in \mathcal{B}} w(B)$ as the optimal base.

\subsection{Uncertainty Matroid}

Traditional exact optimization (P1) aims to find the optimal solution assuming precise knowledge of the true weights $w_e$.  We assume that while the weights $w_e$ associated with $e\in E$ are not precisely known, there is some information available about their potential values.

\begin{definition}[Uncertainty Matroid]
 An uncertainty matroid is a pair $(M,U)$ where $M = (E,I)$ is a matroid and $U$ is a function, mapping $E$ to a nonempty set $U(E)$ $\subseteq$ $\mathbb{R}^n $, the {\em uncertainty area } of $E$.
\end{definition}

The common definitions of uncertainty matroid \cite{merino2019minimum} are in terms of set $U(e)$, for each $e\in E$, and $U(E)$ is viewed as the Cartesian product of $U(e)$'s.  Definition 1 extends uncertainty to the set $E$
 by defining $U(E) \subseteq \mathbb{R}^n$ where $w_e$'range  depends on that of $w_f, f\neq e, f\in E$. Researchers have considered various structural shapes of $U(E)$, such as boxes, ellipsoids, parallelepipeds, and polyhedra, in the context of robust optimization \cite{kara2019stability, ozmen2011rcmars, chassein2017minmax}. 

Our notion of uncertainty centers around treating the weight vector $W=(w_e)_{e\in E}$ as a parametric construct \cite{benabbou2021combining, bertsimas2009constructing}. Specifically, we are provided with $p$ attribute vectors $Y_1,Y_2,\ldots,Y_p \in \mathbb{R}^n$, where $Y_j=(y_{ij})_{1\leq i\leq n}$ for $1\leq j\leq p$ and the weight vector $W$ is defined as a convex combination of these attribute vectors, expressed as $W= \sum_{j=1}^p \lambda_j Y_j$, with $\boldsymbol{\lambda}=(\lambda_j)_{1\leq j\leq p}$, $\boldsymbol{\lambda} \in \mathbb{R}^p$ satisfying $0\leq \lambda_j \leq 1$ and $\sum_{j=1}^p \lambda_j=1$. Thus, $U(E)$ is a parametric polyhedral uncertainty area. 

Our methodology embraces, in a sense, a data-driven approach when modeling the uncertainty associated with the vector $W$. In real-world scenarios, obtaining a precise distribution for $W$ is often unattainable. Instead, we typically possess a finite set of $p$ observations of this uncertain vector. In such circumstances, a data-driven approach becomes the more fitting choice, as it capitalizes on the available realizations of uncertain parameters, which frequently constitute our primary source of information. Bertsimas in \cite{bertsimas2009constructing} profess a similar philosophy and have demonstrated that convex uncertainty sets exhibit various intriguing characteristics in the realm of risk theory, giving rise to coherent risk measures. In a similar vein, \cite{benabbou2021combining}  interpret this model of uncertainty as a form of multi-objective optimization where the $W$ is derived by amalgamating these $p$ objectives in diverse ways, a process guided by the parametrized uncertainty area defined by the parameter vector $\boldsymbol{\lambda} \in R^n$.

\subsection{Optimization in Uncertainty Matroid}

Minimax regret is tailored for robust decision-making under conditions of uncertainty. Instead of pursuing a globally optimal solution, its objective is to minimize the maximum regret that could arise from choosing a suboptimal decision when faced with uncertainty.

A specific selection $W$ from the uncertainty area $U(E)$,  $W^s=(w_e^s)_{e\in E}$  is called a scenario $s$ and $w^s(B)$ denotes the sum of weights of the basis $B$ under scenario $s$. For a scenario $s$, we define $z^*(s)=\min_{B\in \mathcal{B}} w^s(B)$.  For a given scenario $s$, (P1) can be handled efficiently to obtain $z^*(s)$ and $B^*(s)=\arg\min_B w^s(B)$. For a given basis $B$, the regret of $B$ in scenario $s$ is defined as $\text{Regret}(B,s)=w^s(B)-z^*(s)$. We use $W^s \in U(E)$  interchangeably as $s \in U(E)$. The maximum regret of $B$ is defined as $ \text{Max\_Regret}(B)=\max_{s\in U(E)} \text{Regret}(B,s)$.  
The worst-case scenario $s^*(B)$ of $B$ is the scenario that maximizes the regret of $B$. In other words, $ 
s^*(B)=\arg\max_{s\in U(E)} \text{Regret}(B,s).
$
Minimax regret (MMR) is the minimum of $s^*(B)$ over all bases of the matroid and it is defined as
\begin{equation}
\begin{split}
    \text{MMR}(U) &= \min_B \left(\text{Max\_Regret}(B)\right) \\
                    &\quad  = \min_B \left(\max_{s\in U(E)} \text{Regret}(B,s)\right) \\ 
                    &\quad  = \min_B \left(\max_s (w^s(B)-z^*(s))\right).
\end{split}
\end{equation}

It is pertinent to use the concept of uniformly optimal basis here. 

\begin{definition}[Uniformly Optimal Basis]
    A basis $B \subseteq \mathcal{B}$ is a uniformly optimal basis of $(M, U)$  if for every realization $W$ in $U$, $B$ is a minimum weight basis \cite{merino2019}. 
\end{definition}

The existence of uniformly optimal basis implies $MMR=0$.

\subsection{Preference Elicitation}

Elicitation allows decision-makers to gather additional information to reduce uncertainty, making more informed decisions. There are several elicitations, and they broadly fall into two categories, namely \textit{Query Elicitation} and \textit{Preference Elicitation}.  Query elicitation is a process of asking targeted questions or making inquiries to gather information about uncertain elements, such as their true weights or values  ~\cite{megow2017randomization, olston2000offering, goerigk2015robust}.  Preference elicitation \cite{vayanos2020robust, benabbou2021combining, drummond2014, gelain2010, bourdache2019active}, on the other hand, is concerned with the problem of strategically eliciting the preferences of the users through a moderate number of pairwise comparison queries. Rather than seeking precise data, preference elicitation techniques address elicitations like pairwise comparisons, ranking, or rating-based approaches. The goal is to minimize the number of queries while accurately estimating user preferences.   Both modes of elicitation employ two strategies for elicitation \cite{merino2019minimum}, namely, {\em incremental} and {\em offline}. In the incremental case, the elicitation algorithms operate sequentially, querying elements one by one, and they use the information gathered at each step to inform their choice of the next element to query. In contrast, offline situations require algorithms to identify a set of queries in advance that need to be executed simultaneously to accumulate enough information for effective problem-solving.

\section{Minimax Regret with Preference Elicitation}
\label{Sec:Pr}
We are concerned with the optimization problem in uncertainty matroids with parametrized uncertainty areas. We leverage incremental PE, systematically eliciting pairwise preferences to refine the uncertainty area iteratively. Our aim is to achieve an exact optimum ($MMR = 0$) with minimal queries, as well as offer a threshold-based stopping mechanism ($ MMR \leq \tau$) to accommodate practical constraints. 

\subsection{Problem Formulation}

We are given an uncertainty matroid $(M, U)$ and a $n \times p$ matrix set $\mathbf{Y}$ of $p$ vectors $Y_1,Y_2,\ldots,Y_p \in \mathbb{R}^n$. The uncertainty area $U(E) \subseteq \mathbb{R}^n$ is defined in terms of the parameter vector $\boldsymbol{\lambda} \in \mathbb{R}^p$ as,
\[
U(E) = \left\{ W \,\middle|\, W = \mathbf{Y}\boldsymbol{\lambda}, \, \mathbf{0} \leq \boldsymbol{\lambda} \leq \mathbf{1}, \, \boldsymbol{\lambda} \cdot \mathbf{1} = 1 \right\}
\]

The parametric space $\{\boldsymbol{\lambda} \in \mathbb{R}^p: \mathbf{0} \leq \boldsymbol{\lambda} \leq \mathbf{1}, \, \boldsymbol{\lambda} \cdot \mathbf{1} = 1\}$ is a simplex $C$ in $\mathbb{R}^{p-1}$ with extreme points as $\{\mathbf{0},\mathbf{e}^1,\mathbf{e}^2,\ldots,\mathbf{e}^{p-1}\}$ in $\mathbb{R}^{p-1}$, where $\mathbf{e}^j \in \mathbb{R}^{p-1}$ has $1$ at $j$th position and $0$ elsewhere.

For $\boldsymbol{\sigma} \in C$ we get $\boldsymbol{\lambda} \in \mathbb{R}^p$ by following conversion formula and for each $\boldsymbol{\lambda}$.

\[
\lambda_j = \begin{cases}
\sigma_j, & \text{for } 1 \leq j \leq p-1 \\
1 - \sum_{j=1}^{p-1} \sigma_j, & \text{for } j=p
\end{cases}
\]

Let \(\text{Ext}(C)\) be the extreme points of polyhedron \(C\). In the \(r\)th iteration, after incorporating preference constraints, the uncertainty region in \(\sigma\)-space is defined as polyhedron \(C^r\), with \(C (= C^0)\).  We assume that the adjacency relationship among all pairs of extreme points is available or can be incrementally computed. At iteration $0$, each of $p$ extreme points is adjacent to every other extreme point. 
The problem addressed in the present work can be formally stated as follows:\\

\noindent
\textbf{Problem Statement(Pr) } \textit{Given an uncertainty matroid $(M, U)$ with $U$ as a parametric polygonal region, and a threshold $\tau$,  the problem is to elicit user's preference incrementally to refine $U$ which admits an optimal basis $B$ with MMR $\leq \tau$}

\noindent
\textbf{Remark}- Eliciting user preferences is crucial, as the objective is to reduce the level of uncertainty in $U$ based on the user's (albeit imprecise) knowledge of the weights. If we do not consider user preferences, the problem becomes trivial, as $U$ could be refined to a neighborhood of any extreme point, guaranteeing MMR = 0.

\subsection{Previous Work and Our Contribution}
In a prior study \cite{benabbou2021combining}, two incremental preference elicitation techniques for \textbf{Pr} were introduced.  The first is an interactive greedy algorithm that combines preference queries with the gradual construction of an independent set,  aiming for an optimal or near-optimal base. The second is an interactive local search algorithm based on sequences of exchanges to improve solutions. Both methods require computing MMR at each iteration.  The study in \cite{benabbou2021combining} did not explore query determination specifics, indicating potential for improvement.

In contrast, our proposed algorithm takes a fundamentally different approach. It relies on an efficient method of optimization with fixed weights which leads to significant computational efficiency gains. Notably, in scenarios involving large values of $n$ and $p$, our algorithm showcases its ability to solve the problem within a reasonably shorter CPU time frame. In contrast, the previous method failed to provide a solution for $n > 100$ or $p \geq 10$ in a reasonable time (more than an hour of execution on the same machine).

The previous algorithms relied on an external LP Solver routine, CPLEX, at every iteration to the determination of the set of extreme points of the resulting parametric polyhedron. Consequently, the extreme points that needed to be carried over from one iteration to the next were repeatedly generated at each stage. In contrast, our algorithm avoids this redundant step and generates only those extreme points that emerge anew because of the response to a preference query, dropping the infeasible ones. This optimization contributes to the efficiency and speed of our algorithm, reducing unnecessary computational overhead.

\section{Polyhedral Combinatorics Approach}
\label{Sec:PolyCompApproach}
In this section, we systematically develop concepts and present a novel approach to solve \textbf{\textbf{Pr}} by leveraging principles of polyhedral combinatorics and exchange principles of matroids. Our approach consists of several components, which are described in this section. We commence with the initial uncertainty area $C^0$ and iteratively refine it upon obtaining responses to preference queries. The update process for $C^r$ is detailed in Section 4.1. As $C^r$ evolves, the set of extreme points changes, which is addressed in Section 4.2. The adjustment of the adjacency relationship within $\text{Ext}(C^r)$ is discussed in Section 4.3. Finally, the determination of the most strategic preference query is outlined in Section 4.4. We also discuss the computation of an upper bound on the MMR. 
 
\subsection{Updating $C^r$}

 During PE iteration $r$, we start with $C^{r-1}$ and $\text{Ext}(C^{r-1})$ and a query $\text{\em prefq}(e_l, e_k)$ is posed to know user's preference between $e_l$ and $e_k$. If $e_l$ is preferred over $e_k$, then  $w_l \geq w_k$ and hence, $(\mathbf{Y}. \boldsymbol{\lambda})_l \geq (\mathbf{Y}.\boldsymbol{\lambda})_k$ or, $\sum_{j=1}^p y_{jl} \lambda_j \geq \sum_{j=1}^p y_{jk} \lambda_j$. Expressing in term of $\sigma$, we get the inequality
\begin{equation}
\sum_{j=1}^{p-1} (y_{jl}-y_{pl}-y_{jk}+y_{pk}) \sigma_j + y_{pk}-y_{pl} \geq 0
\label{eq:example}
\end{equation}

This is denoted as $\mathbf{a^r} \boldsymbol{\sigma} - b^r \geq 0$, where $a_j^r = (y_{jl}-y_{pl}-y_{jk}+y_{pk})$, $1 \leq j \leq p-1$, and $b^r=(y_{pl}-y_{pk})$ and using this  $C^{r-1}$ is refined to $C^r$ as follows.
\begin{equation}
C^r=C^{r-1}\cap\{\boldsymbol{\sigma}: \mathbf{a^r} \boldsymbol{\sigma} - b^r \geq 0\}
\label{eq:example}
\end{equation}

\subsection{Updating $\text{Ext}(C^r)$}

Imposing the constraint, given as inequality (2), makes some of the points in $\text{Ext}(C^r)$ infeasible and some new extreme point is created. Let $\boldsymbol{\sigma}^u=(\sigma_j^u)_{1\leq j \leq p}$ and $\boldsymbol{\sigma}^v=(\sigma_j^v)_{1\leq j \leq p}$ be two adjacent extreme points in $\text{Ext}(C^{r-1})$ and using (2), the new constraint   $\mathbf{a}^r \boldsymbol{\sigma} - b^r \geq 0$ is such that 
\[
\sum_{j=1}^{p-1} a_j^r \sigma_j^u - b^r > 0 \quad \text{and} \quad \sum_{j=1}^{p-1} a_j^r \sigma_j^v - b^r < 0
\]
The hyperplane $\mathbf{a}^r \boldsymbol{\sigma} - b^r = 0$ intersects the line joining $\boldsymbol{\sigma}^u$ and $\boldsymbol{\sigma}^v$ at a point determined by parameter $\theta$ satisfying the following:
\[
\sum_{j=1}^{p-1} a_j (\theta \sigma_j^u+(1-\theta) \sigma_j^v) - b =0
\]
Simplifying, $\theta$ is determined as
\[
\theta=\frac{b-\sum_{j=1}^{p-1} a_j \sigma_j^v}{\sum_{j=1}^{p-1} (\sigma_j^u-\sigma_j^v) a_j}
\]
With this value of $\theta$, the new extreme point $\sigma^{uv}$ that replaces $\sigma^v$ is computed as
\begin{equation}
\boldsymbol{\sigma}^{uv}=\theta \boldsymbol{\sigma}^u+(1-\theta) \boldsymbol{\sigma}^v
\label{eq:example}
\end{equation}
Thus,
\begin{equation}
\text{Ext}(C^r )\leftarrow\text{Ext}(C^{r-1})\setminus\{\boldsymbol{\sigma}^v\}\cup\{\boldsymbol{\sigma}^{uv}\}
\label{eq:example}
\end{equation}

It is worthwhile to mention that  many extreme points of $C^{r-1}$ will be dropped while some new extreme points will be added. 

\subsection{ Updating the adjacency structure of $\text{Ext}(C^r)$ }
In this section, we identify the rules of adjacency when a new extreme point is generated in an iteration.  A \textit{face} of a convex polytope is any intersection of the polytope with a halfspace such that none of the interior points of the polytope lie on the boundary of the halfspace. Equivalently, a face is the set of points satisfying equality in some valid inequality of the polytope \cite{lovasz2009matching}. If a polytope is \((p-1)\)-dimensional, its facets are its \((p - 2)\)-dimensional faces, its extreme points are its \(0\)-dimensional faces, its edges are its \(1\)-dimensional faces with two extreme points as endpoints, and its \(k\)-dimensional face is defined by its \((k+1)\) extreme points.

\begin{proposition}
For extreme points $\boldsymbol{\sigma}^u$ and $\boldsymbol{\sigma}^v$ if $\boldsymbol{\sigma}^u, \boldsymbol{\sigma}^v \in \text{Ext}(C^{r-1})$ and $\boldsymbol{\sigma}^u, \boldsymbol{\sigma}^v\in C^r$ then $\boldsymbol{\sigma}^u, \boldsymbol{\sigma}^v\in \text{Ext}(C^r)$  and the adjacency relationship between $\boldsymbol{\sigma}^u$ and $\boldsymbol{\sigma}^v$ in $\text{Ext}(C^{r-1})$ is retained in $\text{Ext}(C^r)$ .
\end{proposition}

 The proof is trivial as if $\boldsymbol{\sigma}^u$ and $\boldsymbol{\sigma}^v$  are in 
    $C^r$, then both points satisfy the inequality Eq. 2 and lie on the same side of the hyperplane. Hence the adjacency relation between them is unaffected.

\begin{proposition}
If in iteration $(r-1)$, $\boldsymbol{\sigma}^u, \boldsymbol{\sigma}^v\in \text{Ext}(C^{r-1})$  and are adjacent but  in iteration $r$, $\boldsymbol{\sigma}^u\in \text{Ext}(C^r)$ and $\boldsymbol{\sigma}^v\notin \text{Ext}(C^r)$ then
\begin{itemize}
\item A new extreme point $\boldsymbol{\sigma}^{uv}$ is generated, as in Eq 4,  in place of $\boldsymbol{\sigma}^v$ and 
\item The adjacency relationship between $\boldsymbol{\sigma}^u$ and $\boldsymbol{\sigma}^v$ is carried forward to the pair $\boldsymbol{\sigma}^u$ and $\boldsymbol{\sigma}^{uv}$.
\end{itemize}
\end{proposition}

\begin{proof} \textit{(Sketch)}
     The hyperplane defined by inequality in Eq. 2 intersects the line joining $\boldsymbol{\sigma}^u$ and $\boldsymbol{\sigma}^v$, making \(\boldsymbol{\sigma}^v\) infeasible. If \(\boldsymbol{\sigma}^u\) and \(\boldsymbol{\sigma}^v\) are adjacent in \(C^r\), then the line segment joining them defines a $1$-dimensional face (or edge) of the polytope. The intersection point of this line segment with the hyperplane yields a new extreme point ($0$-dimensional face), \(\boldsymbol{\sigma}^{uv}\) that is adjacent to \(\boldsymbol{\sigma}^u\).
  
 \end{proof}
The following proposition is the converse of the above.
\begin{proposition}
Every new extreme point of $C^r$ is obtained from some pair of extreme points $\boldsymbol{\sigma}^u\in \text{Ext}(C^{r-1}) \cap C^r $ and 
$ \boldsymbol{\sigma}^v\in \text{Ext}(C^{r-1}) \setminus C^r $ and  of the form $\boldsymbol{\sigma}^{uv}$ defined in Eq 4. 
\end{proposition}
\begin{proof} \textit{(Sketch)}
    An extreme point of the polytope is a 0-dimensional face. A new extreme point is generated as the intersection of the hyperplane corresponding to the new constraints with a  1-dimensional face. Every 1-dimensional face is a line joining some pairs of existing extreme points.
\end{proof}

\begin{proposition}
For extreme points $\boldsymbol{\sigma}^w$, $\boldsymbol{\sigma}^u$, and $\boldsymbol{\sigma}^v$ if $\boldsymbol{\sigma}^w$, $\boldsymbol{\sigma}^u$, $\boldsymbol{\sigma}^v \in \text{Ext}(C^{r-1})$ defining a $2$-dimensional face of the polytope and $\boldsymbol{\sigma}^w \in C^r$, $\boldsymbol{\sigma}^u $ and $\boldsymbol{\sigma}^v \notin C^r$ then $\boldsymbol{\sigma}^{wu}$ and $\boldsymbol{\sigma}^{wv}$ are adjacent in $C^r$ if $\boldsymbol{\sigma}^u$ and $\boldsymbol{\sigma}^v$ are adjacent in $C^{(r-1)}$.
\end{proposition}
\begin{proof} \textit{(Sketch)} With similar logic as above, new extreme points are generated due to the intersection of the hyperplane (Eq 4) with 1-dimensional faces. a pair of new extreme points are adjacent only if these are on a 1-dimensional face which results from the intersection of the hyperplane with a 2-dimensional face. 

\end{proof}
Following proposition is a generalization of the above.
\begin{proposition} Assume that the extreme points $\boldsymbol{\sigma}^{u_1}$, $\boldsymbol{\sigma}^{u_2}, \ldots, \boldsymbol{\sigma}^{u_t} \in \text{Ext}(C^{r-1})$ constitute a $2$-dimensional face of the polytope. If  $\boldsymbol{\sigma}^{u_1}$, $\boldsymbol{\sigma}^{u_2}, \ldots, \boldsymbol{\sigma}^{u_k}  \in C^r$, and $\boldsymbol{\sigma}^{u_{k+1}}$, $\boldsymbol{\sigma}^{u_{k+2}}, \ldots, \boldsymbol{\sigma}^{u_t}  \notin C^r$ then the two new extreme points are $\boldsymbol{\sigma}^{u_{1}u_{t}}$ and $\boldsymbol{\sigma}^{u_{k}u_{k+1}}$ and are adjacent in $C^r$.
\end{proposition}
Proof can be derived on the same line as in the previous proposition.

\subsection{Selecting  Strategic Preference Query}

At iteration $r$, we solve P1 at each $\boldsymbol{\sigma}^{j} \in \text{Ext}(C^r)$ with corresponding $\boldsymbol{\lambda}^j$ and  scenario $s_j$.  We get $z^{*s_j} = \min_{B} w^{s_j}(B)$ and $B^{*s_j} = \arg\min_B w^{s_j}(B)$.

\noindent The following observation is straightforward to prove.

\begin{proposition} If there exists a basis $B$ such that $w^{s_j}(B) = z^{*s_j}$ for each $\boldsymbol{\sigma}^j \in \text{Ext}(C^r)$, then $B$ is a uniformly optimal basis in $C^r$. Further, when we have a uniformly optimal solution in $C^r$, then $MMR=0$ in $C^r$.
\end{proposition}

We make use of this condition as the optimality test to get an exact optimal solution. Our objective is to incrementally select a few preference queries, which gradually refine the uncertain area till we reach a uniformly optimal basis. We describe below the process of selecting a preference query at every iteration in the event the optimality test fails.

A {\em preference query} is {\em prefq}($l$, $k$) is a query that seeks revelation of decision maker’s preference between elements $e_l$ and $e_k$ in $E$. The response to {\em prefq}($l$, $k$) is either “$e_l$ is preferred to $e_k$” or “$e_k$ is preferred to $e_l$”.  When there is no uniformly optimal basis, assume that the basis $B^u$ is optimal at  $\boldsymbol{\sigma}^u$ and $B^v$ is optimal at  $\boldsymbol{\sigma}^v$  with $B^u \neq B^v$ and $\boldsymbol{\sigma}^u \neq \boldsymbol{\sigma}^v$ and none of $B^u$ and $B^v$ is optimal at both $\boldsymbol{\sigma}^u$ and  $\boldsymbol{\sigma}^v$. We define {\em Disparity Pair} as a pair of elements $(e_l, e_k) \in E $  in $C^r$ if  $l \in B^u \setminus B^v$ and $k \in B^v \setminus B^u$. 

A pair ($l$, $k$)can be a disparity pair for multiple pairs of $u$ and $v$. We call the {\em frequency} of ($e_l$, $e_k$) as the number of pairs of bases having ($e_l$, $e_k$) as a disparity pair. The disparity pair with the highest frequency is chosen to construct a preference query.  
The underlying intuition behind this process is rooted in the strong exchange axiom of matroids, and this aspect is discussed while justifying the correctness of the proposed algorithm.

\subsection{Upper Bound on MMR}
To conserve computational resources and streamline the analysis, we compute the MMR exclusively for the extreme points to showcase this trend rather than calculating it for the entire uncertainty area. This quantity acts as an upper bound for true MMR. Let $\widetilde{\mathcal{B}}(r) \subseteq \mathcal{B} $ be the collection of optimal bases for problem P1 solved at all extreme points of $C^r$. We compute an approximation of MMR at iteration $r$ as follows.

\begin{equation}
\begin{split}
    \widetilde{\text{MMR}}(U^r) &= \min_{B \in \widetilde{\mathcal{B}}(r)} \left(\text{Max\_Regret}(B)\right) \\
                    &  = \min_{B \in \widetilde{\mathcal{B}}(r)} \left(\max_{s\in \text{Ext}(C^r)} \text{Regret}(B,s)\right). 
\end{split}
\end{equation}

In the following section, we will elucidate our proposed algorithm, which comes in two versions.

\section{Algorithm Pr-A}
\label{Sec:Pr-A}

Algorithm~\ref{Algo:Pr-A} describes the proposed algorithm, {\em Pr-A} to solve \textbf{Pr} based on the components described in the foregoing discussion. 
In a bird's eye view, the algorithm incrementally refines the polyhedral uncertainty area, updates its extreme points by computing only the newly generated ones,  records the adjacency relations, checks if uniformly optimal is achieved, computes an estimate on MMR and generates the most strategic preference query.

\begin{algorithm}[ht!]
    \SetAlgoLined
    \SetAlgoLined
    \SetKwData{Left}{left}\SetKwData{This}{this}\SetKwData{Up}{up}
    \SetKwFunction{Union}{Union}\SetKwFunction{FindCompress}{FindCompress}
    \SetKwInOut{Input}{Input}\SetKwInOut{Output}{Output} \SetKwInOut{Initialization}{Initialization}
    \Initialization{$C^0 = \text{Convex hull}(0, e^1, e^2, \ldots, e^{(p-1)})$, \\ $\text{Ext}(C^0) = \{0, e^1, e^2, \ldots, e^{(p-1)}\}$, \\ $\text{Adjacency}(C^0)$: All pairs in  $\text{Ext}(C^0)$, \\$r = 0$} 
    \While{$MMR > \tau$}{
        $r \leftarrow r + 1$\;
        
        \textbf{Traditional Greedy Step:} For each $s \in \text{Ext}(C^{(r-1)})$, solve P1 by greedy algorithm\;
        Compute $\widetilde{\mathcal{B}}$\;
        
        \textbf{Optimality Check:} If any $B^*$ is uniformly optimal, return $B^*$. \textbf{STOP}\;
        
        \textbf{Compute $\widetilde{MMR}$:} Compute an upper bound of MMR as in section 4.5\;
        
        \textbf{Generate Preference Query:} Find the most frequent disparity pair $(l, k)$, as in Sec 4.4\;
        
        \textbf{Get Response:} Is $l$ preferred to $k$?\;
        
        \textbf{Update $C^r$:} As described in Sec 4.1, Eq 3\;
        
        \textbf{Update Ext($C^r$):} As per Sec 4.2, Eq 5\;

        \textbf{Update Adjacency($C^r$):}  as per the rules given in Sec 4.3
    }
    \caption{Pr-A}
    \label{Algo:Pr-A}
\end{algorithm}

\subsection{Correctness of Pr-A}
The correctness of Pr-A is hinged primarily on the exchange axiom of matroids. The bases of a matroid $M$ on a finite set of elements $E$ satisfy the following exchange axiom:

\noindent
{\bf Exchange Axiom}: For all $ B, B' \in \mathcal{B} $ and each $e' \in B' \setminus B$,  there exists some $e \in B \setminus B'$ such that  $B \setminus \{e\} \cup \{e'\} \in \mathcal{B}$.

The exchange axiom has a mirror version, the co-exchange axiom \cite{berczi2024}:

\noindent
{\bf Co-Exchange Axiom}: For any $ B,B' \in \mathcal{B} $ and $ e \in B \setminus B' $ there exists $e' \in B' \setminus B$ such that $B \setminus \{e\} \cup \{e'\} \in \mathcal{B}$.

Both versions imply the existence of mutually exchangeable elements $e \in B$ and $e'\in B'$ between two bases $B$ and $B'$.  The exchange axiom implies the existence of a sequence of exchanges that transforms $B$ into $B'$, and the number of exchange steps needed to transform $B$ to $B'$ is $|B \setminus B'|$. Any exchange pair is a disparity pair between $B$ and $B'$, but a disparity pair is not necessarily an exchange pair. Let the degree of disparity between any two bases $B$ and $B'$ be the number of disparity pairs between them. 

There is another intriguing observation in this context. Let us consider two scenarios $s$ and $s'$, corresponding to two extreme points $\boldsymbol{\sigma}$ and $\boldsymbol{\sigma}'$ of $C^r$. The line segment joining $\boldsymbol{\sigma}$ and $\boldsymbol{\sigma}'$ is denoted as $\text{line}[\boldsymbol{\sigma}, \boldsymbol{\sigma}']$. Suppose that $B^s$ and $B^{s'}$ are the optimal bases of $\boldsymbol{\sigma}$ and $\boldsymbol{\sigma}'$, respectively and that the set of optimal bases along this line as $B[\text{line}[\boldsymbol{\sigma}, \boldsymbol{\sigma}']] = \{ B^{s''} \}_{s'' \in \text{line}[\boldsymbol{\sigma}, \boldsymbol{\sigma}']}$. The set of bases in $B[\text{line}[\boldsymbol{\sigma}, \boldsymbol{\sigma}']]$ contains the bases corresponding to the pairwise exchanges to transform $B^s$ to $B^{s'}$. As one moves along $B[\text{line}[\boldsymbol{\sigma}, \boldsymbol{\sigma}']]$ from $\boldsymbol{\sigma}$ to $\boldsymbol{\sigma}'$, the number exchange pairs decrease, and the degree of disparity is non-increasing.  In simpler terms, for any $B^{s''}$ in $\text{line}[\boldsymbol{\sigma}, \boldsymbol{\sigma}']$, the degree of disparity between $B$ and $B''$ is not more (possibly, less) than that between $B$ and $B'$. 

This suggests that when the line segment $\text{line}[\boldsymbol{\sigma}, \boldsymbol{\sigma}']$ is divided at an intermediate step (say, at a non-extreme point $\boldsymbol{\sigma}''$ with $B''$ as an optimal base) into two smaller segments, the degree of disparity between $B$ and $B''$, as well as that between $B'$ and $B''$, are either less than or equal to the degree of disparity between $B$ and $B'$. If $(u, v)$ forms a valid disparity pair between $B$ and $B'$, then introducing an inequality corresponding to a preference query $(u, v)$ divides $\text{line}[\boldsymbol{\sigma}, \boldsymbol{\sigma}']$ into two smaller segments (one feasible and the other infeasible) and both disparity does not increase in any of the sub-segments.  

\begin{figure}[htbp]
    \centering
    \begin{subfigure}[b]{0.33\textwidth}
        \includegraphics[width=2in, height=2.05in]{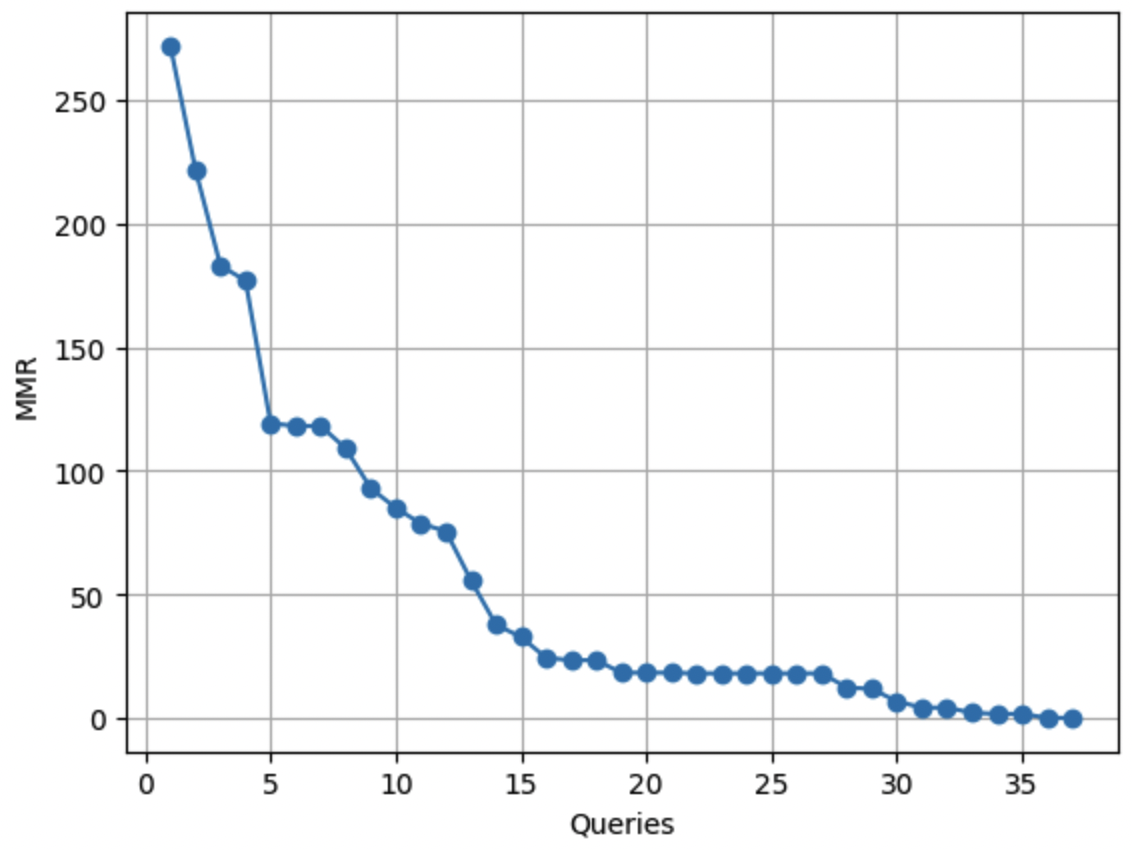}
        \caption{}
        \label{fig:mmr6}
    \end{subfigure}%
    \begin{subfigure}[b]{0.33\textwidth}
        \includegraphics[width=2in, height=2in]{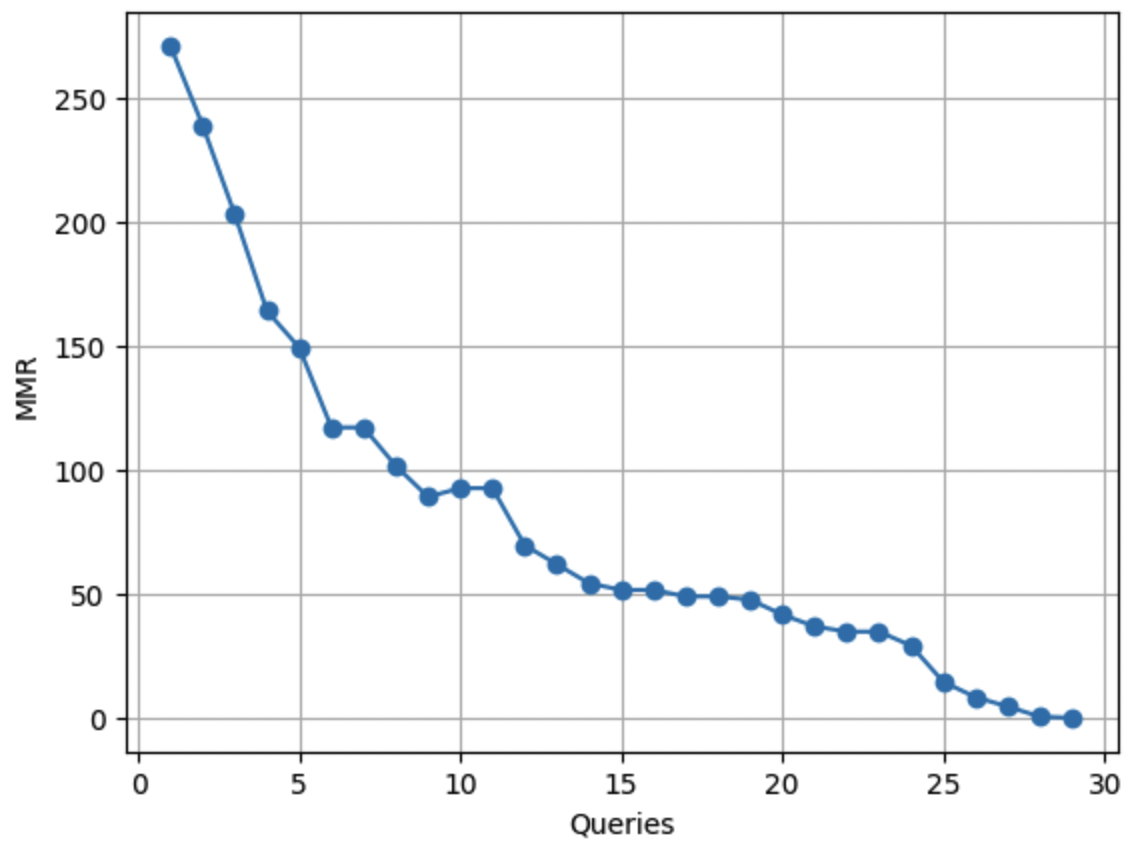}
        \caption{}
        \label{fig:mmr7}
    \end{subfigure}%
    \begin{subfigure}[b]{0.33\textwidth}
        \includegraphics[width=2in, height=2in]{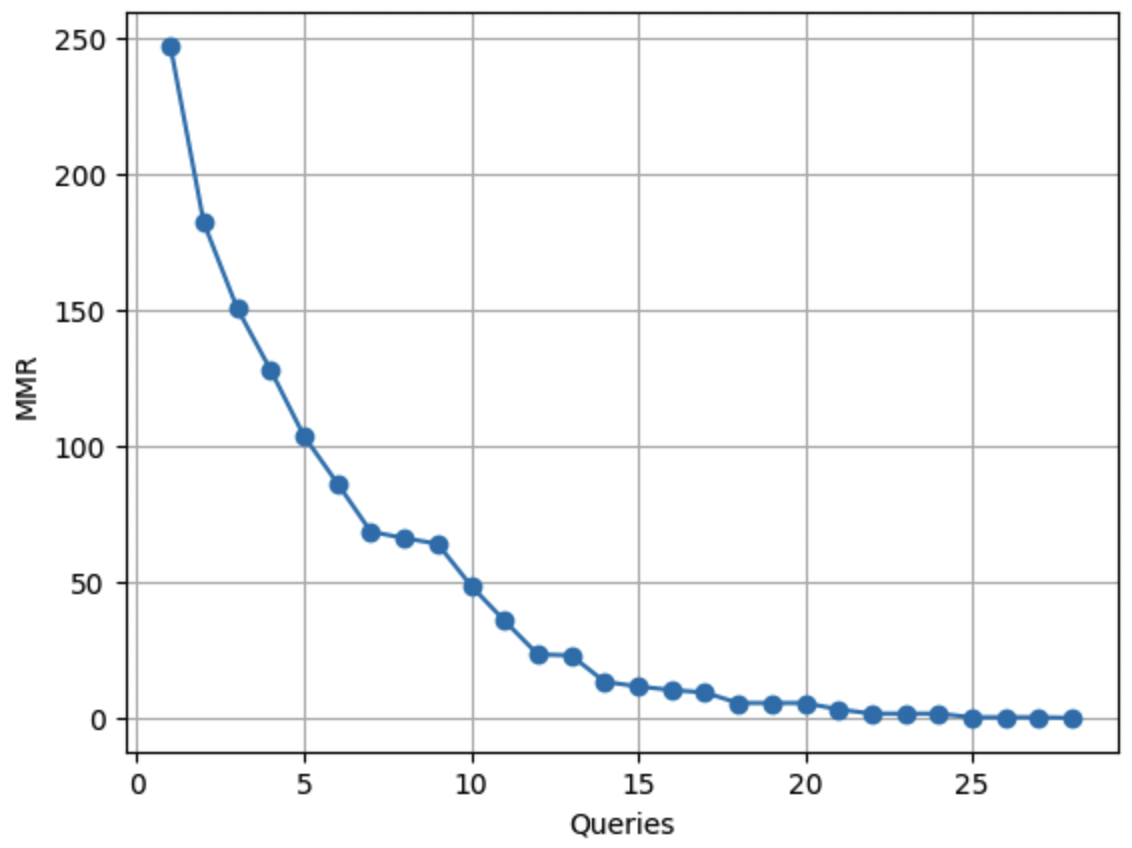}
        \caption{}
        \label{fig:mmr5}
    \end{subfigure}
    \begin{subfigure}[b]{0.33\textwidth}
        \includegraphics[width=2in, height=2.05in]{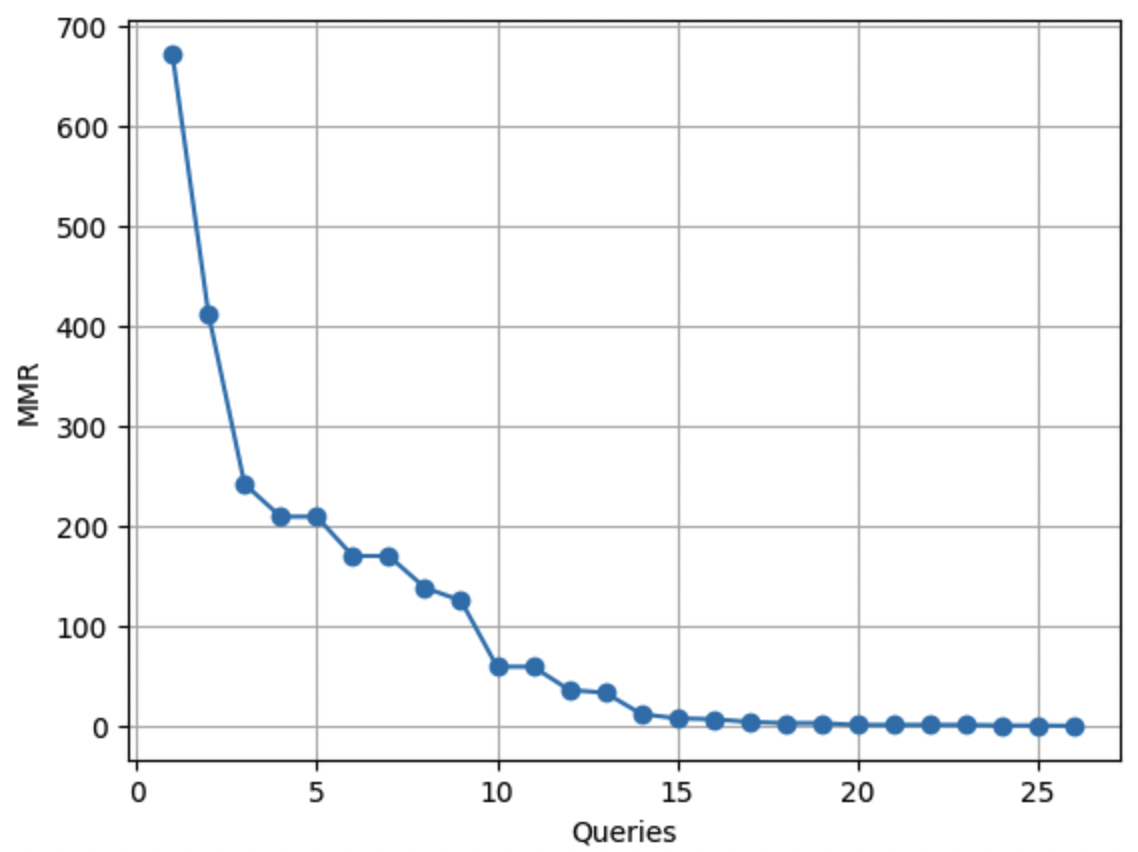}
        \caption{}
        \label{fig:mmr4}
    \end{subfigure}%
    \begin{subfigure}[b]{0.33\textwidth}
        \includegraphics[width=2in, height=2in]{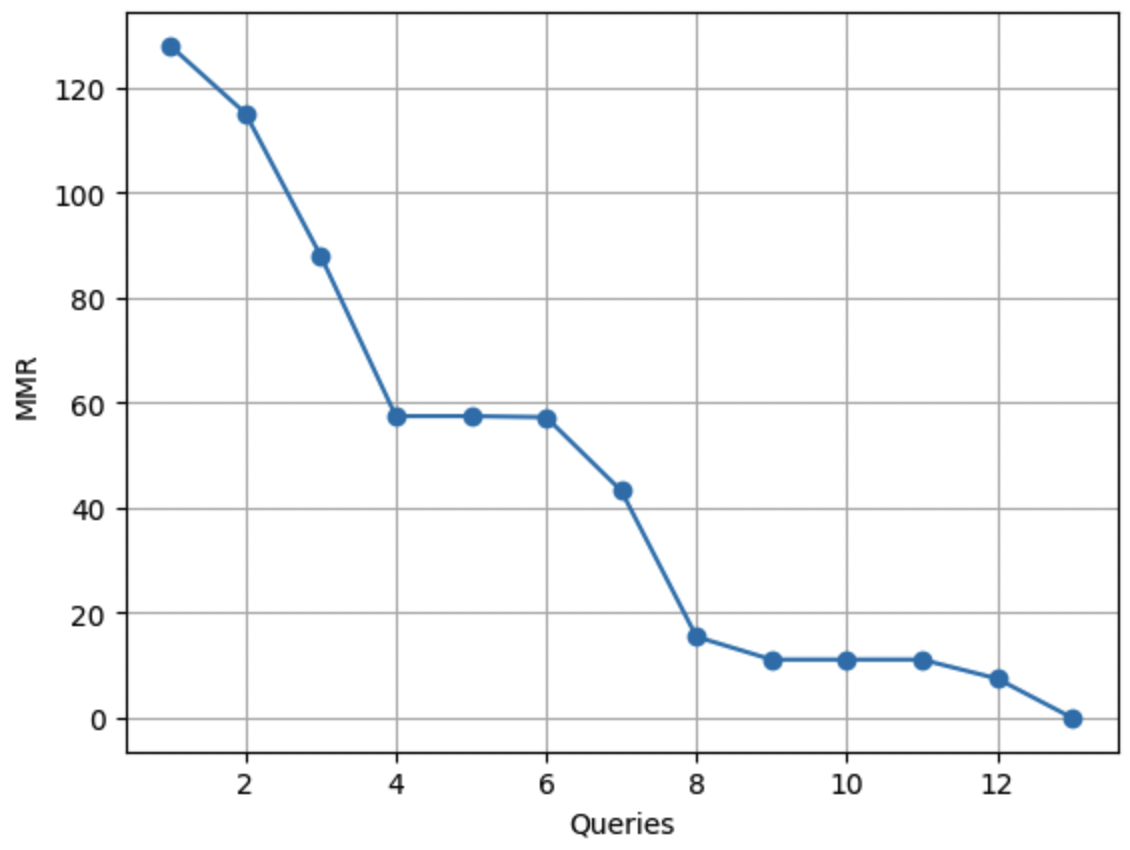}
        \caption{}
        \label{fig:mmr3}
    \end{subfigure}
    \caption{Rate of Convergence of Pr-A with varying values of $n$ and $p$.  (a) $n =10, ~p=10$. (b) $n =20, ~p=10$. (c) $n =30, ~p=10$. (d) $n =40, ~p=10$. (e) $n =50, ~p=10$.}
    \label{fig:rate_of_convergence}
\end{figure}

The preceding observation suggests that when preference queries are selected based on disparity, the degree of disparity gradually approaches zero. As a result, the MMR also converges. While there are numerous options for choosing preference queries based on disparity, we employ a strategy of choosing the preference query corresponding to the most frequently occurring disparity pair. While this heuristic doesn't guarantee the best choice, intuitively, the most frequent disparity pair involves the maximum number of pairs of $\mathcal{B}$, thereby increasing the probability of rapid convergence. This observation is also confirmed in our experimental analysis (Figure \ref{fig:mmr_trend} and Figure \ref{fig:rate_of_convergence})

\subsection{Illustration of working of Pr-A}
 We illustrate the working of the algorithm using a toy problem. 

A scheduling matroid with 8 jobs is considered with deadlines as $(5, 3, 2, 4, 1, 5, 3, 2)$ units of time. The uncertainty area is defined by 4 parameter vectors $Y_1$, $Y_2$, $Y_3$, and $Y_4$, and the matrix $\mathbf{Y}$ is given by
\[
\renewcommand*{\arraystretch}{1.1}
\mathbf{Y} = 
\begin{bmatrix}
6  & 8  & 8  & 3  \\
2  & 4  & 7  & 7  \\
5  & 2  & 5  & 6  \\
8  & 7  & 1  & 8  \\
1  & 2  & 8  & 2  \\
6  & 3  & 3  & 7  \\
3  & 4  & 6  & 5  \\
2  & 3  & 1  & 4  \\
\end{bmatrix}
\]

The uncertainty area $C^0$ is the convex hull of $4$ extreme points as $0$, $\mathbf{e}^1$, $\mathbf{e}^2$, and $\mathbf{e}^3$. When the greedy step is employed at each of these extreme points, we get the optimal bases at each of $4$  extreme points are   $\{1, 3, 4, 6, 7 \}$, $\{1, 2, 4, 6, 7 \}$,  $\{1, 2, 5, 6, 7 \}$,  and $\{2, 3, 4, 6, 7 \}$. For notational convenience, job $i$ is denoted as $i$ here.

Consequently, the disparity pairs are $(1, 2)$, $(1, 3)$, $(1, 4)$, $(2, 3)$, $(2, 4)$, $(3, 5)$, and $(4, 5)$ and the most frequently occurring disparity pair is $(4, 5)$. As a result, the preference query aims to elicit the preference between job $4$ and job $5$. It is assumed that job $4$ is preferred over job $5$, leading to the inequality constraint $ \lambda_1 - \lambda_2 - 13 \lambda_3 +6 \geq 0$ to obtain $C^1$. The set of extreme points $Ext(C^1)$ consists of 6 points: $e^1$, $e^2$, $e^4$, and the coordinates $(0.5, 0, 0.5)$, $(0, 0.5833, 0.4167)$, and $(0, 0, 0.4615)$. The next iteration commences where we compute the most frequent disparity pair as $(5, 6)$.  We take that job $6$ is preferred to job $5$, and a new inequality is added to get $C^2$, which has $7$ extreme points. The algorithm runs for $8$ iterations and terminates with $MMR=0$. Other preference queries are $(5, 7),(3, 5), (2, 7), (3, 7), (4, 7), (1, 7)$.

When two extreme points possess distinct optimal bases, there exists a base that represents the optimal solution along the line connecting these extreme points. 
This base is the outcome of an exchange process involving the optimal bases associated with these extreme points. Additionally, at this juncture, one of the original bases becomes an optimal basis as well. Therefore, during this process, the algorithm effectively increases the number of extreme points that share a common optimal base. Intuitively, we work towards achieving optimality conditions or towards the uniformly optimal base. Notably, in this process, the minimax regret (MMR) monotonically decreases. We empirically demonstrate this phenomenon in Figure~\ref{fig:mmr_trend}.

\section{Experimental Results}
\label{Sec:Exp}
In this section, we conduct a thorough analysis of our experimental results to reinforce the concepts outlined earlier in this paper. 
\begin{figure}[ht!]
    \centering
    \begin{subfigure}[b]{0.48\textwidth}
        \includegraphics[width=3in, height=2.5in]{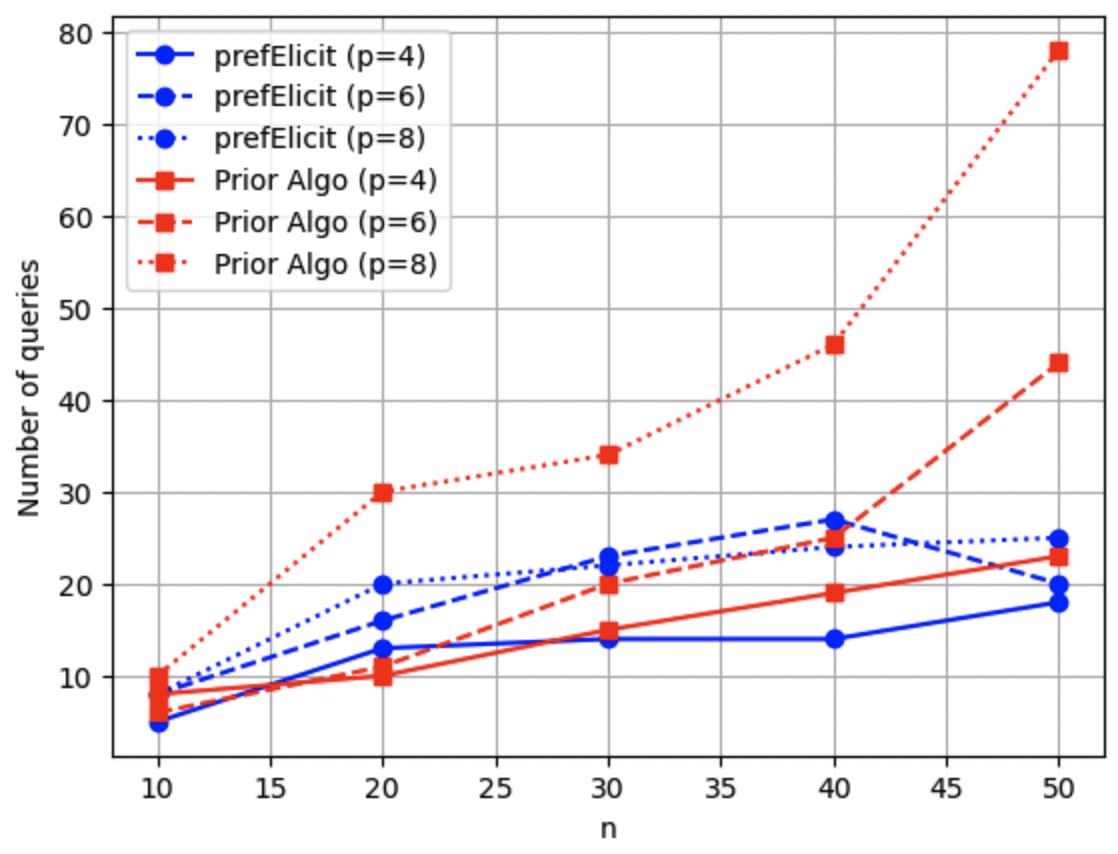}
        \caption{Scheduling Matroid}
        \label{fig:sq}
    \end{subfigure}%
    \begin{subfigure}[b]{0.48\textwidth}
        \includegraphics[width=3in, height=2.5in]{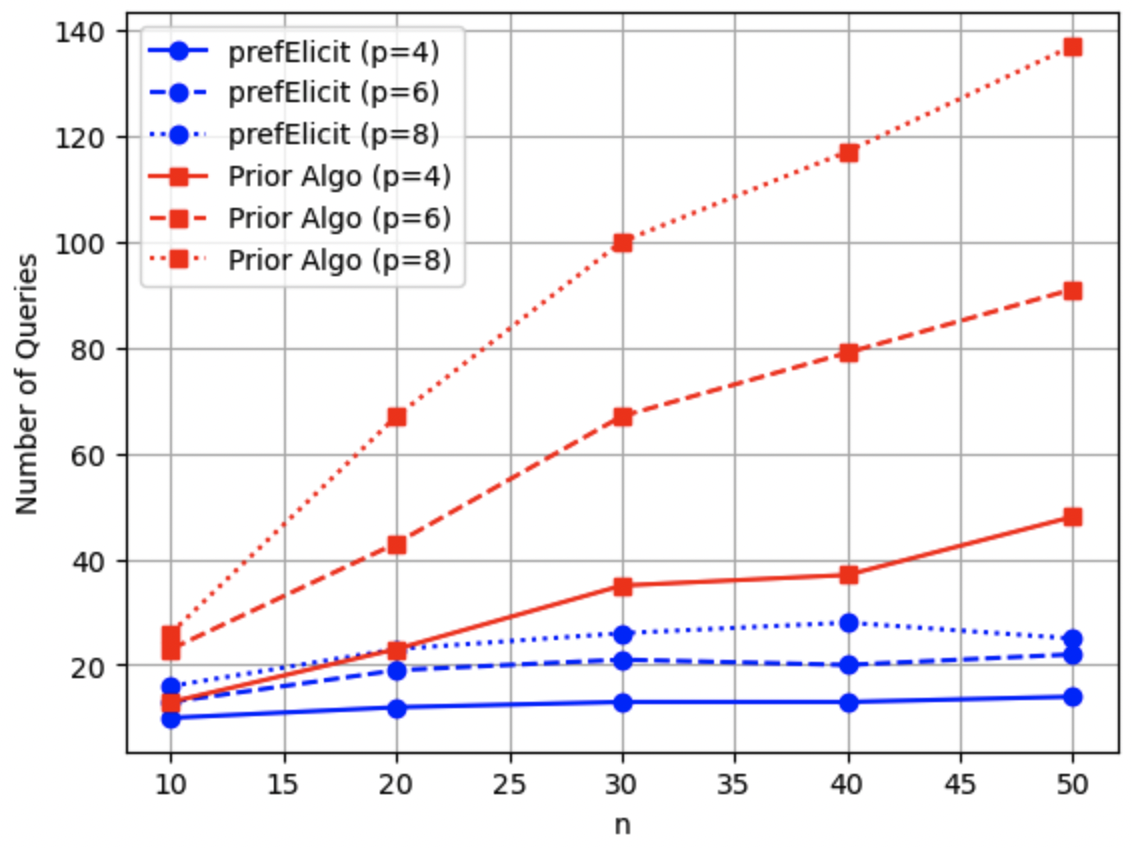}
        \caption{Graphic Matroid}
        \label{fig:gq}
    \end{subfigure}
    \begin{subfigure}[b]{0.48\textwidth}
        \includegraphics[width=3in, height=2.5in]{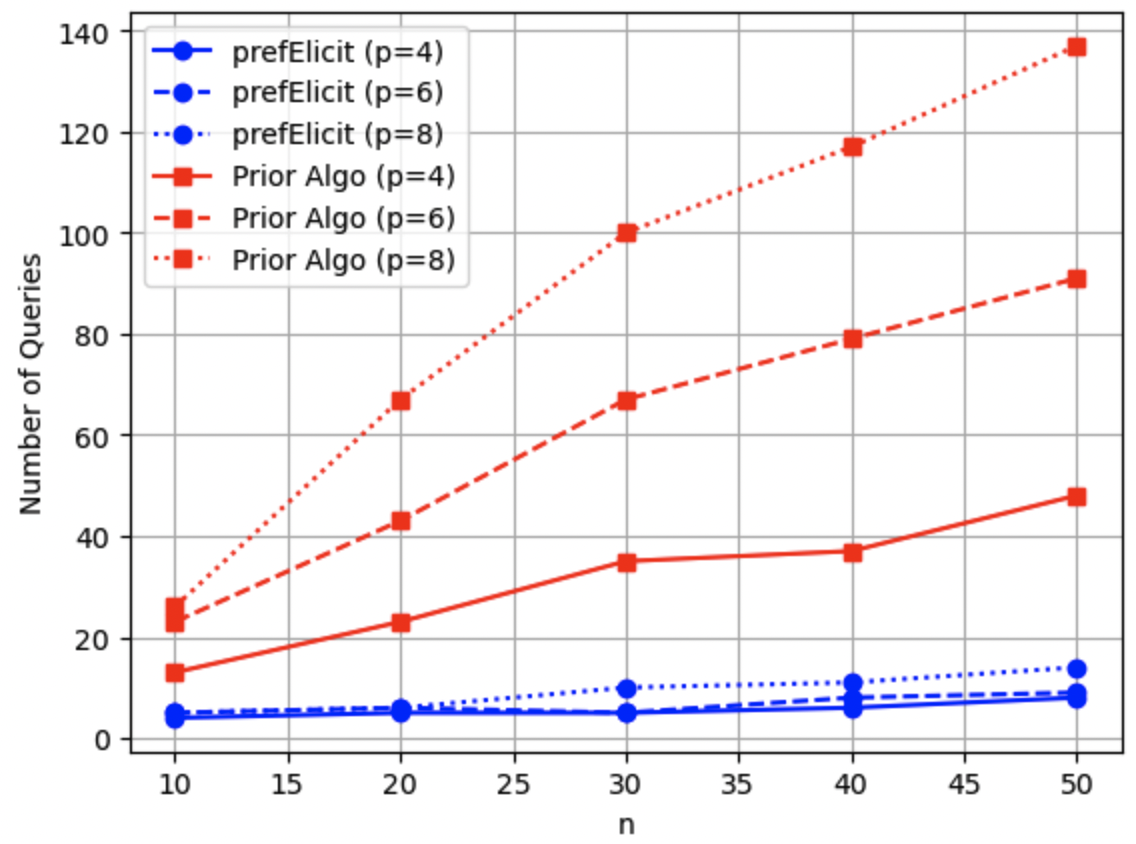}
        \caption{Uniform Matroid}
        \label{fig:uq}
    \end{subfigure}%
    \begin{subfigure}[b]{0.48\textwidth}
        \includegraphics[width=3in, height=2.5in]{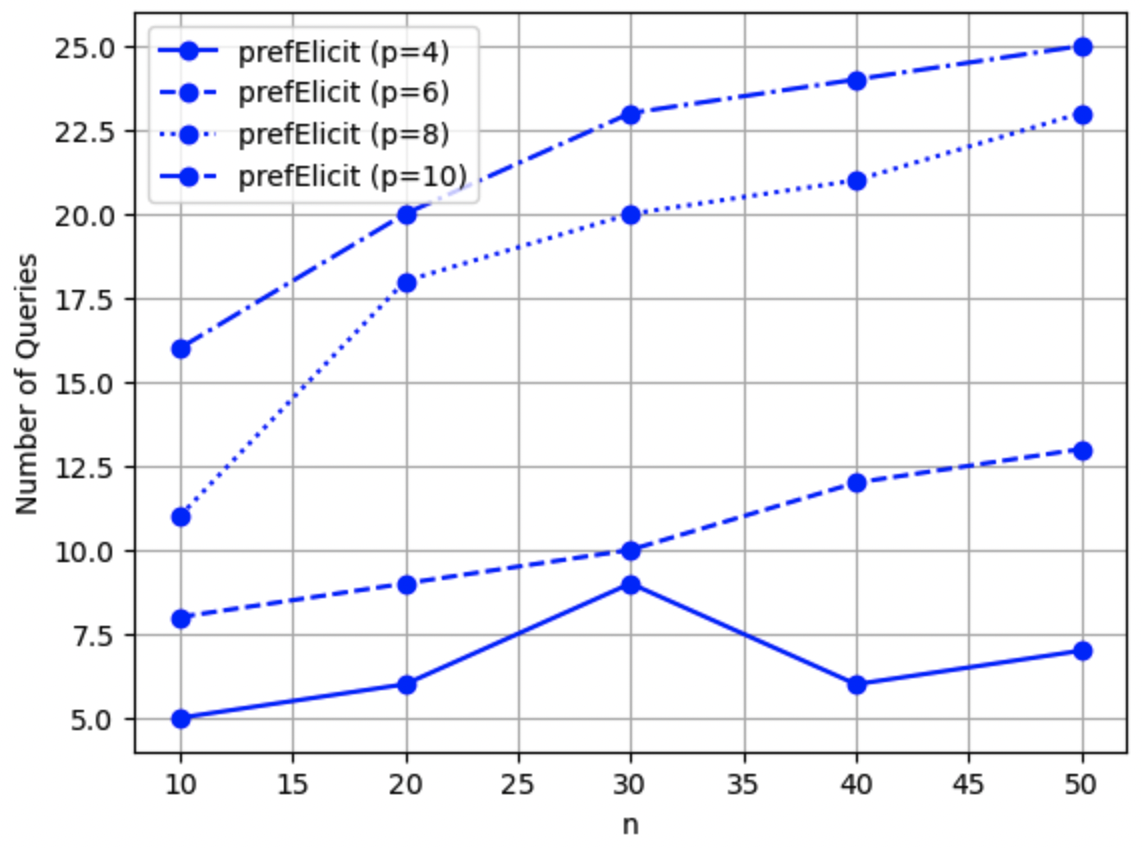}
        \caption{Partition Matroid}
        \label{fig:pq}
    \end{subfigure}
    \caption{Performance comparison for various matroids: (a) scheduling matroids, (b) graphic matroids, (c) uniform matroids, (d) partition Matroid.}
    \label{fig:performance}
\end{figure}
Our main objective is to showcase the convergence of MMR and validate the efficiency of our method in terms of reduced computational time and preference queries. We provide empirical support for our heuristic approach, emphasizing the strategic selection of preference queries based on the most frequent disparity pair. Additionally, we investigate the trade-off in scenarios where MMR exceeds zero, highlighting significant time and effort savings. 
\begin{figure}[ht!]
    \centering
    \begin{subfigure}[b]{0.5\textwidth}
        \includegraphics[width=3in, height=2.5in]{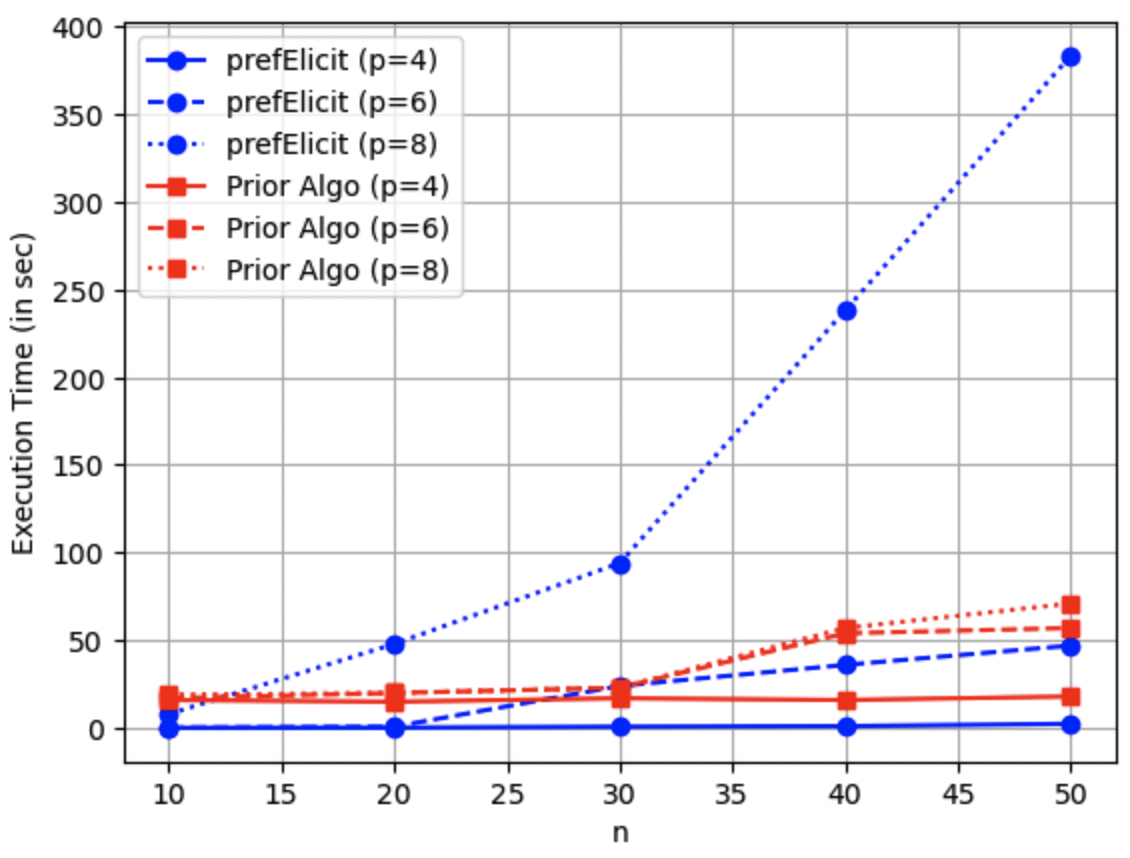}
        \caption{Scheduling Matroid}
        \label{fig:st}
    \end{subfigure}%
    \begin{subfigure}[b]{0.5\textwidth}
        \includegraphics[width=3in, height=2.5in]{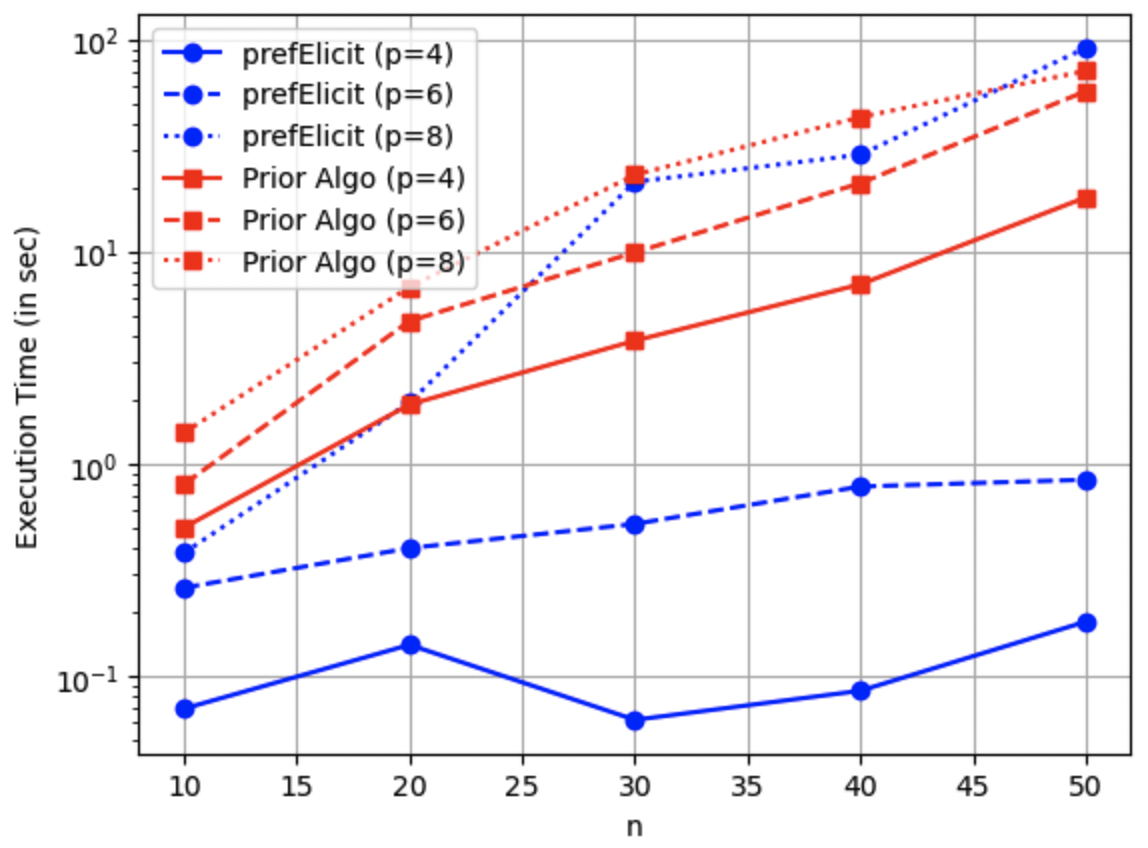}
        \caption{Graphic Matroid}
        \label{fig:gt}
    \end{subfigure}
    \begin{subfigure}[b]{0.5\textwidth}
        \includegraphics[width=3in, height=2.5in]{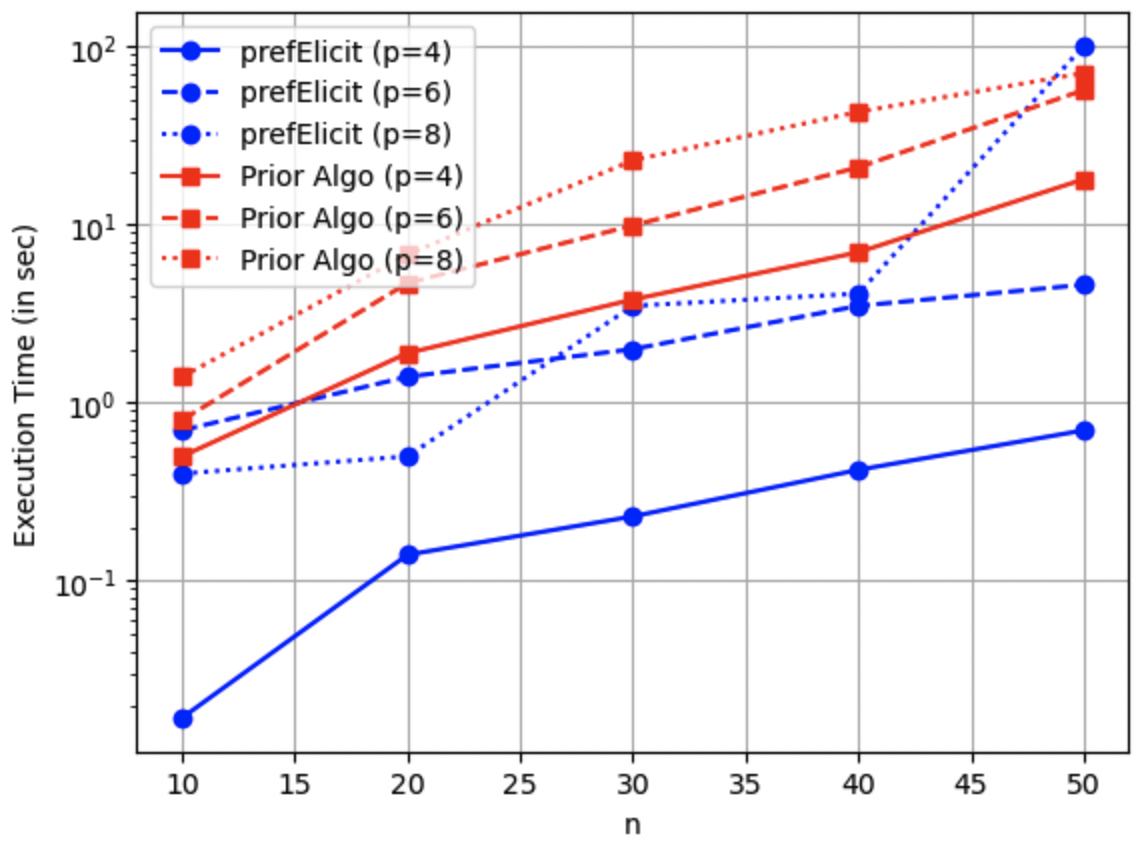}
        \caption{Uniform Matroid}
        \label{fig:ut}
    \end{subfigure}%
    \begin{subfigure}[b]{0.5\textwidth}
        \includegraphics[width=3in, height=2.5in]{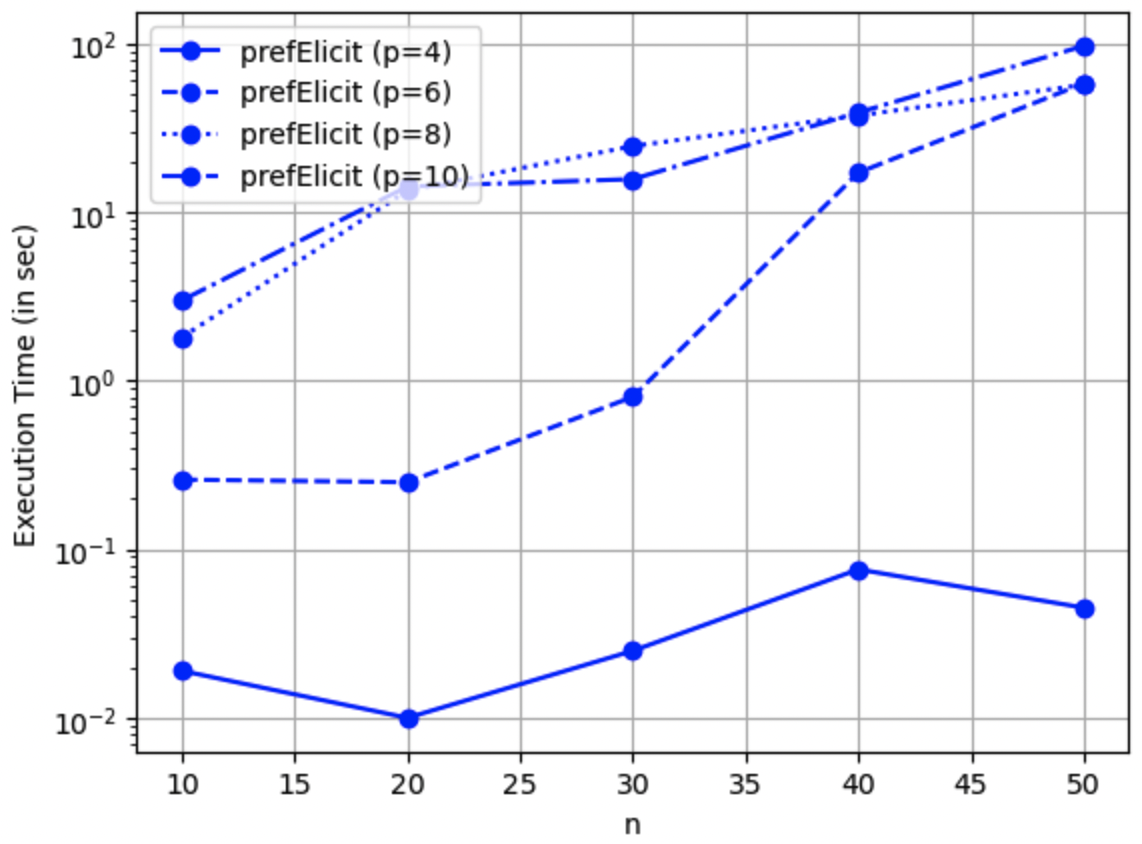}
        \caption{Partition Matroid}
        \label{fig:pt}
    \end{subfigure}
    \caption{Execution time comparison for various matroids: (a) scheduling matroids, (b) graphic matroids, (c) uniform matroids, (d) partition Matroid.}
    \label{fig:execution_time}
\end{figure}
Our experiments involve the random generation of various matroids with differing dimensions and parameter vectors ($n$ and $p$ values). We implement Pr-As on four distinct types of matroids: uniform, graphic, scheduling, and partition matroids. While the first three matroid types undergo a rigorous comparison between Pr-A, implemented in C, and existing algorithms, for partition matroids, we solely demonstrate the performance of Pr-A. These algorithms are executed under similar conditions on a state-of-the-art desktop system. Performance evaluation is based on the number of preference queries and computation times (expressed in seconds). We conduct experiments with randomly generated instances, averaging the results of 20 runs for each matroid type and combination of $n$ and $p$.

The experimental analysis is presented in Figures 1-7, where some plots depict single instances to illustrate trends observed across all instances. 
In Figure \ref{fig:rate_of_convergence}, we report the performance of Pr-A for various values of $n$ with $p=10$- (a) for $n=10$ (b) for $n=20$ (c) for $n=30$ (d) for $n=40$ and (e) for n=50. These subfigures demonstrate the number of queries and MMR values over iterations. Notably, earlier algorithms require high computational time when $p=10$.

Figures in \ref{fig:performance}  offer a performance comparison regarding the number of queries required by our method and a previous approach \cite{benabbou2021combining} across three distinct matroids, with variations in $n$ and $p$ values. We further present experimental results for partition matroids. The $x$-axis ranges from 10 to 50, representing $n$ values, while the $y$-axis displays the count of preference queries needed to achieve $MMR=0$ for both algorithms. There are four subfigures, each utilizing different scales for the y-axis to aid visualization, corresponding to $p$ values of 4, 6, and 8.  Across all $n$ values and for all four matroids, Pr-A consistently required approximately 25 queries, whereas the earlier algorithm demanded over 100 queries for higher $p$ values, particularly evident when $p =8$. Remarkably, our proposed method demonstrates a substantial reduction in the number of queries needed to achieve a robust optimal solution.

Figure \ref{fig:execution_time} presents a performance comparison regarding the execution time of both algorithms, using the same set of instances as reported in Figure \ref{fig:performance}. Notably, for the scheduling algorithm, both methods perform competitively, except in cases where $p=8$, where the proposed algorithm exhibits higher execution time. Particularly, for $p=10$, both algorithms run for significantly longer duration. However, for graphic matroids and uniform matroids, Pr-A demonstrates promising efficiency in terms of computational time. It is observed that as $p$ increases beyond 10, the geometric dimension of the polyhedral region also increases, and hence, none of the algorithms converge to $MMR=0$ in reasonable time. 

\begin{figure}[htbp]
    \centering
    \includegraphics[width=3.5in, height = 2.5in]{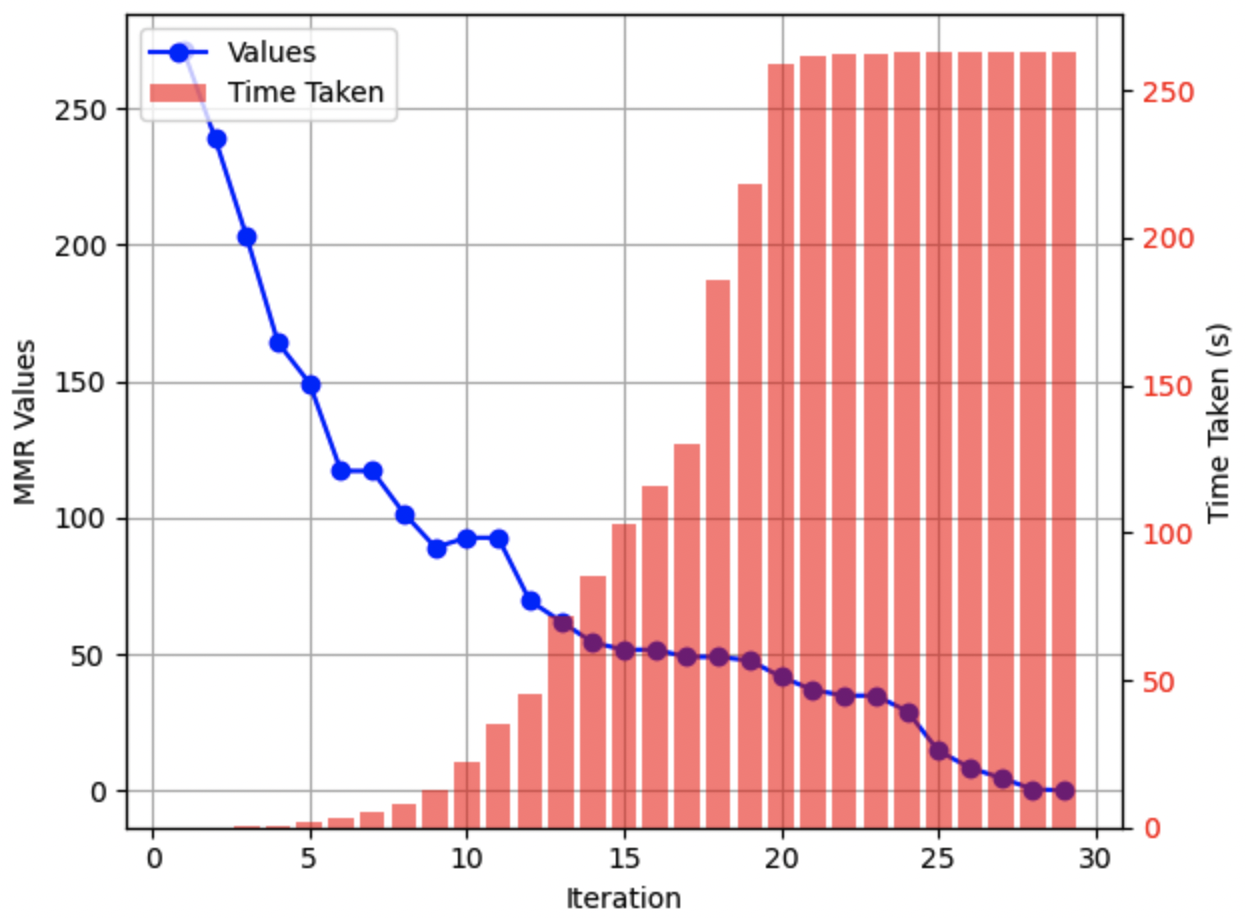}
    \caption{Trend of MMR Variation and Execution Time }
    \label{fig:mmr_trend}
\end{figure}

Figure \ref{fig:mmr_trend} illustrates the trends of MMR variations over iterations for scheduling matroid when $p=10$ and incorporates the corresponding execution time (in seconds). This plot serves to showcase the convergence of Pr-A, depicting the gradual decrease in MMR across iterations. Additionally, it juxtaposes the associated execution time for each iteration, thereby emphasizing the efficiency and effectiveness of our method in achieving an optimal solution. The results underscore that as the most frequent disparity pairs are progressively revealed, the MMR consistently decreases, ultimately converging to zero.

\begin{figure}[ht!]
    \centering
        \includegraphics[width=3.2in, height=2.5in]{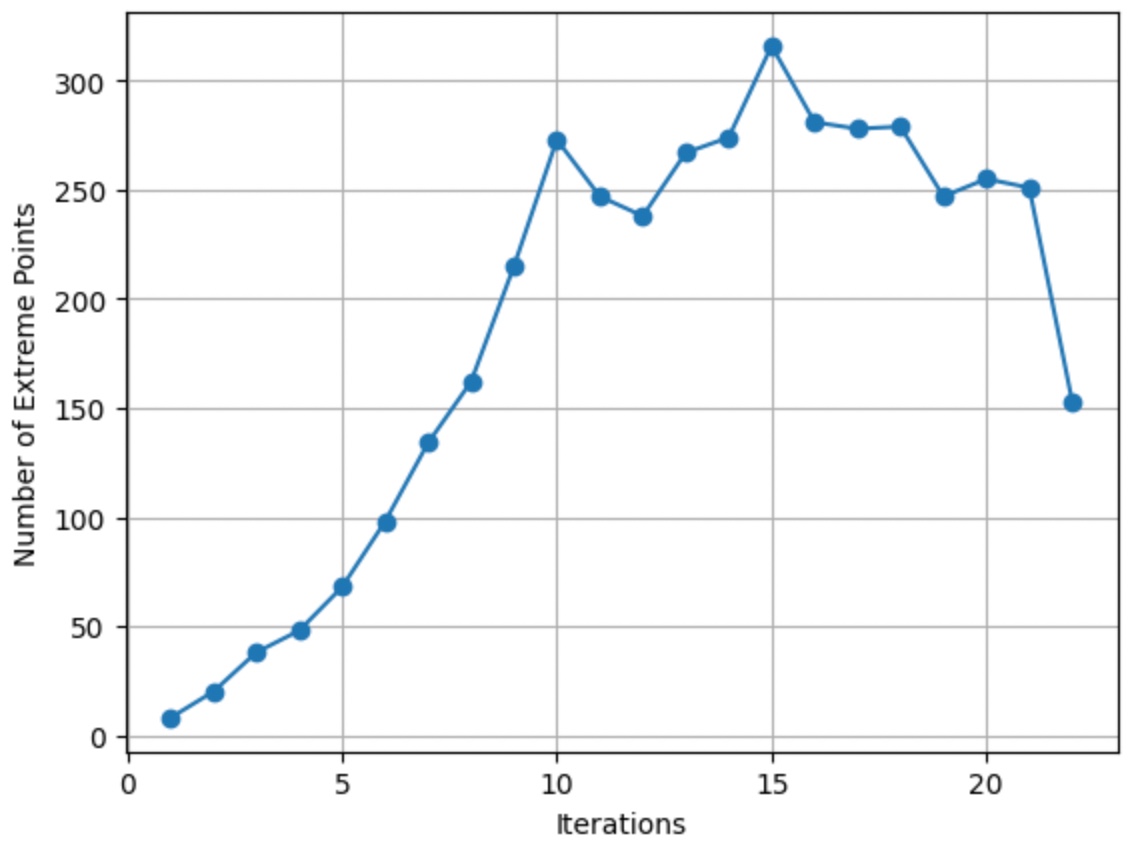}
        \caption{Number of Extreme Points Generated over Iterations}
        \label{fig:mmr2}
\end{figure}

\begin{figure}[ht!]
    \centering
        \includegraphics[width=3.2in, height=2.5in]{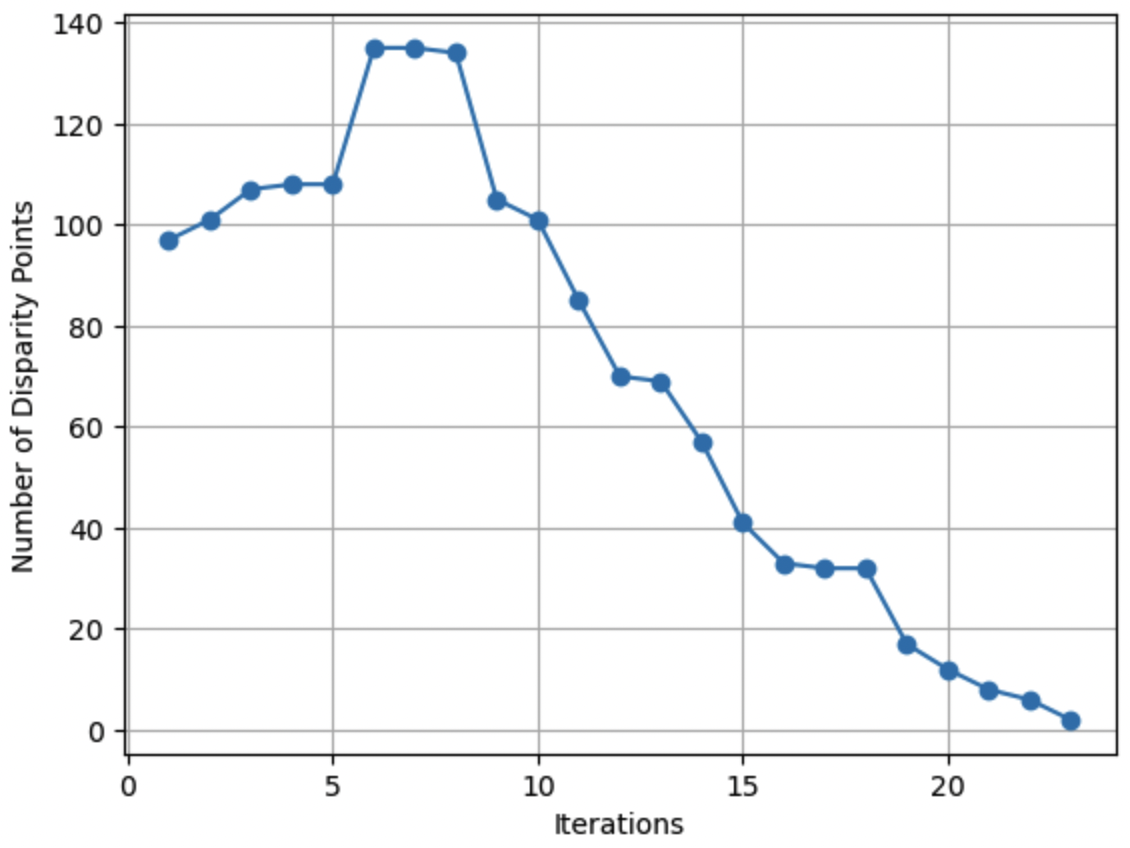}
        \caption{Disparity vs. iterations}
        \label{fig:disparity_points}
\end{figure}

\begin{figure*}
    \centering
    \begin{subfigure}[b]{0.5\textwidth}
    \includegraphics[width=3in, height=2.5in]{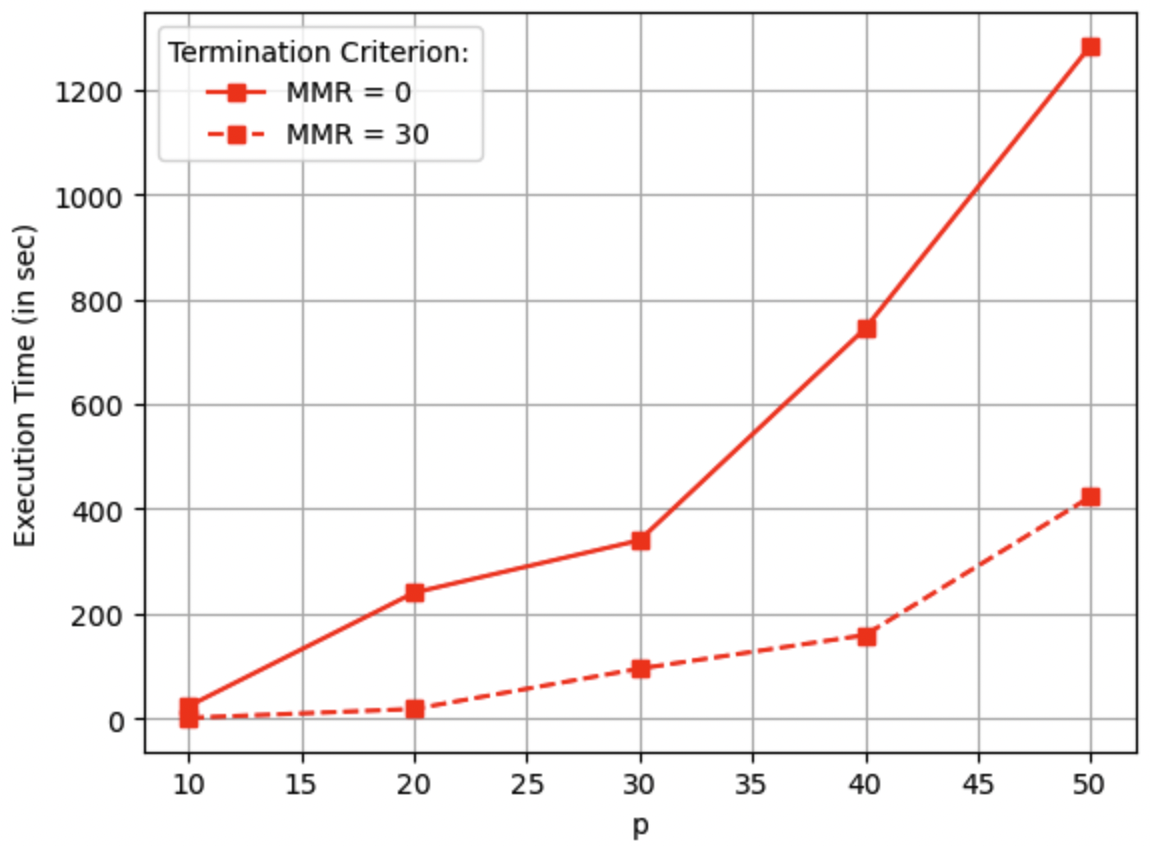}
    \caption{Execution Time vs. Number of Jobs}
    \label{fig:mmrt}
    \end{subfigure}%
    \begin{subfigure}[b]{0.5\textwidth}
    \includegraphics[width=3in, height=2.5in]{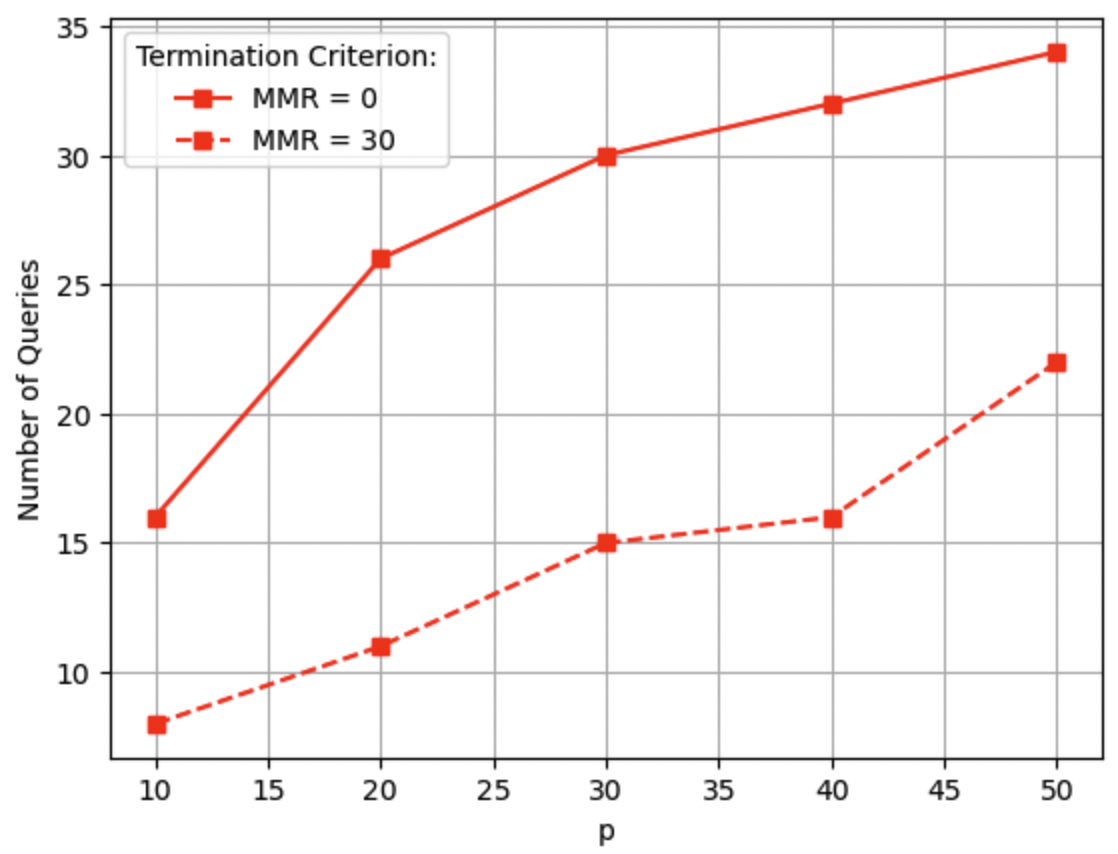}
    \caption{Number of queries vs. Number of Jobs}
    \label{fig:mmrq}
    \end{subfigure}
    \caption{Performance of Pr-A for scheduling matroid for $p=10$ in two scenarios:$MMR = 0$ and $MMR \leq 30$ }
    \label{fig:mmrqt}
\end{figure*}

Figure \ref{fig:mmr2} analyses the number of extreme points generated over iterations. Figure \ref{fig:disparity_points} demonstrates that the number of disparities gradually decreases until convergence. 
This observation serves to justify that as we incorporate more preference queries, the disparity among optimal bases diminishes. This reduction implies that MMR converges to zero. This observation also justified that disparity-based heuristics guarantee convergence.  When we impose  $MMR \leq \tau$ instead of insisting on $MMR=0$, PefElicit demonstrates encouraging results. In this context, Figure \ref{fig:mmrqt} illustrates the advantages of a suboptimal solution with $MMR \leq \tau$ over a robust optimal solution with $MMR=0$. We conducted an experiment with scheduling matroid to measure (a) the execution time and (b) the number of queries varying $n$ (the number of jobs) for both scenarios. Figure \ref{fig:mmrqt} illustrates the performance of our algorithm for two scenarios: MMR=0 and MMR=30. We demonstrate that when $\tau \neq 0$, we may obtain a suboptimal point that is close to the optimal solution. At the same time, we significantly reduce computational efforts, particularly in terms of the number of preference queries required. We conclude that there is a substantial gain in computational efforts (and in the number of queries) if the suboptimal option is chosen.

\section{Conclusions}
\label{Sec:Con}
In this paper, we introduce a novel preference elicitation scheme for minimax regret optimization of uncertainty matroids. Leveraging the parametric polyhedral structure of the uncertainty region, we depart from numerical iteration and adopt a geometric approach to enhance efficiency. Our proposal is grounded in theory and empirically demonstrates superior performance compared to existing techniques. The algorithm offers a method for selecting the most suitable preference query based on the most frequent disparity pair. While it is feasible to generate preference queries based on frequent exchange pairs, this approach demands more computational effort. We intend to explore this line of research in the future.

\bibliographystyle{unsrt}
\bibliography{mybibfile}

\end{document}